\documentclass[10pt,twocolumn,letterpaper]{article}

\usepackage[pagenumbers]{cvpr}

\usepackage{booktabs}
\usepackage{graphicx}
\usepackage{xspace}
\usepackage{amsmath}
\newcommand{\vig}{\mathrm{VIG}}
\newcommand{\vigtot}{\vig_{\mathrm{total}}}
\newcommand{\viglay}{\vig_{\mathrm{layout}}}
\newcommand{\vigglob}{\vig_{\mathrm{global}}}
\newcommand{\Lreal}{\mathcal{L}_{\mathrm{real}}}
\newcommand{\Lperm}{\mathcal{L}_{\mathrm{perm}}}
\newcommand{\Lblank}{\mathcal{L}_{\mathrm{unif}}}

\newcommand{\suppEstimator}{Appendix~\ref{supp:estimator}}
\newcommand{\suppSchedule}{Appendix~\ref{supp:schedule}}
\newcommand{\suppAnswers}{Appendix~\ref{supp:answers}}
\newcommand{\suppGrowth}{Appendix~\ref{supp:growth}}
\newcommand{\suppDigitize}{Appendix~\ref{supp:digitize}}
\newcommand{\suppTraining}{Appendix~\ref{supp:training}}

\newcommand{\suppPublic}{Appendix~\ref{supp:public}}

\definecolor{cvprblue}{rgb}{0.21,0.49,0.74}
\usepackage[pagebackref,breaklinks,colorlinks,allcolors=cvprblue]{hyperref}

\def\confName{CVPR}
\def\confYear{2027}

\title{What Does the Visual Encoder Buy?\\Visual Information Gain in Encoder-Free Multimodal Pretraining}

\author{Zhixuan Ge \qquad Anqi Li \qquad Wei Qiu\textsuperscript{\dag}\\
Rice University\\
{\tt\small \{zg33, al318, wq8\}@rice.edu}\\
{\small \textsuperscript{\dag}Corresponding author}}

\begin{document}
\maketitle
\begin{abstract}
Recent scaling studies ask whether vision-language models still need a visual encoder by comparing encoder-free and encoder-based models on the next-token loss of the text in image-text data. This loss can fall without any use of the image, and two runs with the same loss can differ in what they take from it. We therefore measure what the image contributes. Visual information gain (VIG) is the loss increase when the image is replaced by a uniform one. It splits exactly into global information, carried by image statistics, and layout information, which shuffling the patches within the image destroys. The measurement takes three forward passes, needs no labels, and applies to every checkpoint of a training run and to released models. We apply it to decoders with 158M to 1.04B parameters that read pixel patches through a linear embedding, evaluated every billion tokens up to 40B, and to 19 public checkpoints. In our decoders the image lowers the loss by 3\% to 46\%, depending on the image type. Global information is present after the first billion tokens. Layout information starts near zero, grows slowly with training and does not grow with model size on five of six image types, and a larger batch reaches the same loss with less of it. With a frozen SigLIP~2 encoder in front of the same decoder the loss is lower only when the arrangement of the image is intact. On shuffled images the losses with and without the encoder agree to within 0.04 nats on five of six image types, so the net advantage of the encoder in the loss is layout information, above all the ability to read text in images. On the digits in answers about documents, the image raises accuracy from 62\% to 90\% with the encoder and by 2 to 3 points without it. Twice the OCR supervision and 16-pixel patches do not supply this, nor does a randomly initialized copy of the encoder architecture over its first 8B tokens, so what the encoder contributes in this phase is its pretrained prior. Public encoder-free models at production scale (Gemma~4 12B, NEO, Mono-InternVL) do acquire layout information. On the same documents they have 0.53 to 0.76 nats of it, against 0.65 to 0.82 for recent encoder-based models and 0.04 to 0.07 for our decoders, and their released training stages show it rising. These results agree with the reported parity of the two families when both are trained from scratch and place the reported advantage of a pretrained encoder in layout information.
\end{abstract}

\section{Introduction}
\label{sec:intro}

\begin{figure}[t]
  \centering
  \includegraphics[width=\linewidth]{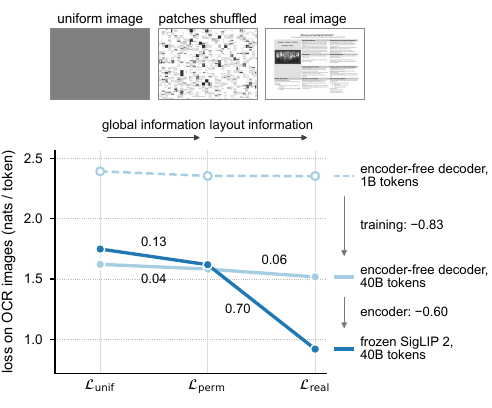}
  \caption{\textbf{What the encoder buys is layout.} The loss of a 323M decoder on OCR images with the image replaced by a uniform one ($\Lblank$), with its patches shuffled ($\Lperm$) and with the real image ($\Lreal$). The step from $\Lblank$ to $\Lperm$ is global information, and the step from $\Lperm$ to $\Lreal$ is layout information. From 1B to 40B training tokens the loss of the encoder-free decoder falls by 0.83 nats while the two steps reach only 0.04 and 0.06, so the descent is language. With a frozen SigLIP~2 encoder the loss on shuffled images is that of the encoder-free decoder and the loss on real images is 0.60 nats lower.}
  \label{fig:teaser}
\end{figure}

Encoder-free vision-language models drop the pretrained vision encoder and feed pixel patches to the language decoder through a lightweight patch embedding~\cite{fuyu2023,chen2024solo,diao2024eve,diao2025evev2,luo2025monointernvl,lei2025sail,diao2025neo}. One transformer then handles both modalities, any resolution and any interleaving of images and text, and no separate vision pretraining is needed. Production models now take this route~\cite{gemma2026gemma4}, and whether the visual encoder is still needed has become a scaling question. Shukor \etal~\cite{shukor2025scaling} train models with and without a separate encoder from scratch and find no inherent advantage of the encoder. Chen \etal~\cite{chen2026howfar} compare encoder-free models with models built on a pretrained encoder and find that the encoder-free models start behind and improve faster with compute. Both studies, like other scaling studies of decoders that read images~\cite{wu2026native}, compare the two families on the next-token loss of the text in image-text data, the multimodal loss.

The multimodal loss is a blunt instrument for a question about vision. A decoder lowers it by modeling the language of captions, questions and answers and by using the image, and the loss does not say which. Three observations from our experiments show what this hides. On document images the loss of an encoder-free decoder falls from 2.35 to 1.52 nats between 1B and 40B training tokens while replacing the image by a uniform one never raises it by more than 0.11 nats, so the descent is language (\cref{fig:teaser}). A decoder trained with a four times larger batch reaches the same loss, 1.433 against 1.432 nats, and depends less on the arrangement of the image on every image type. And the losses of released models cannot be compared with one another, since tokenizers and training text differ, while the change of a loss under a change of the image can. We therefore measure what the image contributes. What does a visual encoder supply, what does an encoder-free decoder have to acquire on its own, and when does it do so?

For every checkpoint we score the same text under the real image, under the image with its patches shuffled, and under a uniform image (\cref{fig:teaser}). The loss increase from real to uniform is the \emph{visual information gain} (VIG). It splits exactly into \emph{global information}, which survives the shuffle and is carried by image statistics, and \emph{layout information}, which the shuffle destroys and which includes reading. The measurement needs no labels, probes or benchmarks, only three forward passes on a fixed validation set, so it runs on every checkpoint of a training run and on released models. We check it against the accuracy of answers, other block sizes of the shuffle and other reference images.

We use VIG at the two ends of the encoder question. At the small end we train a ladder of encoder-free decoders from 158M to 1.04B parameters for 40B tokens with the linear patch embedding of Chen \etal~\cite{chen2026howfar}, the same decoder with a frozen SigLIP~2 encoder~\cite{tschannen2025siglip2} at matched data and batch, and interventions on supervision, patch size, batch size and architecture. We evaluate VIG every billion tokens on six image types. Chen \etal observe a slow early phase before the loss of their models drops sharply and hypothesize that the decoder spends it bootstrapping its visual representations. VIG gives this phase a measurable content. We call the stretch of training in which an encoder-free decoder has far less layout information than models with a pretrained encoder the bootstrap phase, and all our encoder-free runs are in it. At the large end we apply VIG to 19 public checkpoints, 14 of them from three encoder-free families at production scale, where interventions are out of reach and the measurement is not. Three findings follow.
\begin{itemize}
\item \textbf{In the bootstrap phase the loss is mostly language, and layout information comes late.} The real image lowers the loss of the encoder-free decoders by 3\% (GUI) to 46\% (charts) relative to a uniform image. Global information is present after one billion tokens. Layout information starts near zero and grows with training, and on five of six image types it does not grow with model size. A larger batch reaches the same loss with less layout information.
\item \textbf{What the encoder buys is layout information, and it comes with the pretrained weights.} With shuffled images a frozen SigLIP~2 encoder and the linear embedding reach the same loss, to within 0.04 nats on five of six image types, so the net loss advantage of the encoder equals its lead in layout information. The loss gives the size of the advantage and VIG gives its content. On OCR images the frozen-encoder condition has 0.70 nats of layout information at 40B tokens against 0.06 without the encoder, and on the digits that have to be read off a document it gains 27.5 points of answer accuracy from the image, against 1.7 to 3.4 points without it. Three cheaper routes do not raise layout information on OCR images above the baseline: twice the OCR supervision, 16-pixel patches and the SigLIP~2 architecture trained from random weights, over the 8B tokens of its run.
\item \textbf{At production scale encoder-free models acquire layout information.} On the same OCR images, the public encoder-free Gemma~4 12B, NEO and Mono-InternVL have 0.53 to 0.76 nats of layout information. This is an order of magnitude above our decoders and overlaps the range of recent encoder-based models (0.65 to 0.82). Their released training stages show the layout information rising. Model size does not raise it in our ladder, which leaves a pretrained language model, more image-text data and staged training as candidates for closing the gap.
\end{itemize}

These results help explain the two scaling studies (\cref{sec:scaling}). Over the 8B tokens of our run, the encoder architecture without its pretrained weights extracts no more image information than the linear embedding, consistent with the parity of the two families when both are trained from scratch~\cite{shukor2025scaling}. And the order in which image types catch up in Chen \etal~\cite{chen2026howfar} follows the size of the layout gap.

\section{Related work}
\label{sec:related}

\paragraph{Encoder-free vision-language models.}
Fuyu~\cite{fuyu2023} and SOLO~\cite{chen2024solo} feed linearly projected pixel patches to a decoder. EVE~\cite{diao2024eve}, HoVLE~\cite{tao2025hovle} and VoRA~\cite{wang2025vora} align or distil their visual features from a pretrained encoder during training. EVEv2~\cite{diao2025evev2} and Mono-InternVL~\cite{luo2025monointernvl} give the two modalities separate weights inside one transformer, SAIL~\cite{lei2025sail} and NEO~\cite{diao2025neo} study single transformers trained on pixels at scale, and NaViL~\cite{tian2025navil} adds a newly initialized vision encoder to a pretrained language model and trains both jointly. Gemma~4~\cite{gemma2026gemma4} and PixelUMM~\cite{pixelumm2026} are recent encoder-free production and unified models, and Tuna-2~\cite{liu2026tuna2} reports that its encoder-based variant converges faster early in pretraining and is overtaken at scale. EVE reports slow convergence and large performance gaps~\cite{diao2024eve}, and Mono-InternVL remains behind its encoder-based counterpart on high-resolution document benchmarks~\cite{luo2025monointernvl}. We quantify where the lag is and test three candidate remedies. Chen \etal~\cite{chen2026howfar} and Harwath \etal~\cite{harwath2026implicit} show that the shallow layers of encoder-free models take on the role of an encoder. We measure what this implicit encoder delivers to the loss.

\paragraph{Scaling laws for multimodal pretraining.}
Compute-optimal scaling~\cite{kaplan2020scaling,hoffmann2022chinchilla} has been extended to mixed-modal generative models~\cite{aghajanyan2023scaling}, to models with and without a separate vision encoder trained from scratch~\cite{shukor2025scaling}, to native multimodal mixture-of-experts pretraining~\cite{wu2026native}, to joint language and visual objectives~\cite{tong2026beyond} and to the comparison of encoder-free and encoder-based pretraining~\cite{chen2026howfar}. The laws for decoders that read images~\cite{shukor2025scaling,wu2026native,chen2026howfar} are fitted to the loss on text tokens. Smooth loss curves can hide stepwise learning of individual tokens~\cite{stepwise2026}, and the level of the pretraining loss is a weak indicator of cross-modal alignment~\cite{huang2024mir}. We decompose the multimodal loss into an image-blind loss and two visual parts, and track each.

\paragraph{Measuring reliance on the image.}
The loss or perplexity difference between text scored with and without its image has been used to filter interleaved documents~\cite{interns2026preview} and to select instruction-tuning samples~\cite{maharana2025adapt}. Language priors in visual question answering (VQA) were exposed with balanced image pairs and shifted answer priors~\cite{goyal2017making,agrawal2018dont}, and blind baselines show that many benchmark questions can be answered without the image~\cite{chen2024mmstar}. Cross-modal interaction has been tested by projecting it out~\cite{hessel2020does}, by ablating one modality~\cite{frank2021vision} and by Shapley attribution~\cite{parcalabescu2023mmshap,hasic2026which}. Permuting the visual tokens after the encoder probes whether tuned models use their order~\cite{lee2024intriguing,qi2025beyond}, shuffled image patches give a structure-disrupted view for contrastive decoding~\cite{xia2026sdcd}, and blank or mismatched images probe grounding~\cite{zafar2026beyond}. Vision transformers are more robust to patch permutation than convolutional networks~\cite{naseer2021intriguing}, and contrastive models behave like bags of words~\cite{yuksekgonul2023bags}. Language models acquire visual priors from text alone~\cite{seebeforeseeing2025}, and such priors lie in the image-blind loss that our measurement separates out. Encoder comparisons usually rely on downstream benchmarks after instruction tuning~\cite{karamcheti2024prismatic,tong2024cambrian}. VIG differs from these measures in three ways. The uniform image keeps the token positions of the image, so the image-blind loss is measured with the sequence intact. The shuffle splits the gain into global and layout information. And the measurement is applied to the pretraining loss of every checkpoint, so it tracks what is learned from the image during pretraining and not what a tuned model uses at test time.

\section{Measuring visual information gain}
\label{sec:method}

\subsection{Definition}
\label{sec:vig}

The multimodal loss is the mean next-token loss in nats on the text tokens of a set of image-text sequences. We score the same text under three image conditions (\cref{fig:teaser}): the real image ($\Lreal$), the shuffled image, in which the $32\times32$ pixel patches are permuted uniformly at random within the image ($\Lperm$), and a uniform image, gray and of the same size, which occupies the same token positions ($\Lblank$). The visual information gain and its two parts are
\begin{align}
  \vigtot &= \Lblank - \Lreal, \label{eq:vig}\\
  \vigglob &= \Lblank - \Lperm, \label{eq:vigglob}\\
  \viglay &= \Lperm - \Lreal, \label{eq:viglay}
\end{align}
so that $\vigtot=\vigglob+\viglay$ holds exactly. $\Lblank$ is the image-blind loss, what the model reaches when the image carries no information. The shuffle keeps every patch, and with it the color histogram, the texture statistics and the shapes of the characters it contains, and destroys their arrangement. $\vigglob$, which we call global information, is therefore what the model extracts from the bag of patches, and $\viglay$, layout information, is what it extracts from their arrangement. Reading text in an image needs the characters in their order, so reading counts as layout information. We define all three as loss differences on the same text so that they compare directly with the loss. A fall of the loss with the real image is the fall of the image-blind loss plus the growth of $\vigtot$.

\paragraph{Estimator.}
We pack 32 sequences of 4{,}096 tokens per image type from held-out data, with 96 (OCR) to 609 (captions) images. The sequences and the shuffle of each image are fixed for every checkpoint and run. The six image types are captions, charts, general visual question answering, OCR images of documents, STEM figures and GUI screenshots (\suppEstimator). The multimodal validation loss that we quote for a run is the mean over the six types on the first 8 of the 32 sequences per type, evaluated during training. Checkpoint noise is largest on captions and charts, up to 0.03 nats for $\viglay$ and 0.05 for $\vigglob$, and below 0.01 on the other types (\suppEstimator). End-of-run values are means over the last three checkpoints.

\paragraph{Checks of the measure.}
Three checks support this reading (\suppEstimator). A uniform image is an input that no model has seen in training. With the image of another sample of the same type as reference, the loss of the frozen-encoder condition on OCR images lies 0.89 nats above the loss with the real image and that of the encoder-free decoder of the same size 0.09, against 0.82 and 0.10 with the uniform image. The order of the two front ends holds with random pixels as reference and with shuffle blocks of 64 and 128 pixels. And the accuracy on answer tokens read off the image orders the models as $\viglay$ does (\cref{sec:encoder}).

\subsection{Models and training}
\label{sec:setup}

\paragraph{Encoder-free decoder.}
As in Chen \etal~\cite{chen2026howfar}, each $32\times32$ pixel patch is flattened, embedded by a linear layer and given two-dimensional position embeddings. We call this front end the linear embedding. The decoder is a dense pre-norm transformer over the Qwen2.5 vocabulary~\cite{qwen2025qwen25}. Attention is bidirectional among the tokens of one image and causal elsewhere, with document masking~\cite{dong2024flex}, and the loss is computed on text tokens only. We train four sizes with 158M, 323M, 633M and 1.04B non-embedding parameters, with batches of 128, 256, 256 and 512 sequences of 4{,}096 tokens (\suppTraining). Chen \etal train mixture-of-experts decoders with 1.1B to 44B total parameters. A dense ladder at our sizes lets us evaluate VIG every billion tokens and repeat runs under interventions.

\paragraph{Data.}
Every batch holds text and image-text tokens in a 1:1 ratio. The image-text half draws from 19.8M samples (6.2B tokens per epoch) of CC12M and CC3M captions~\cite{changpinyo2021cc12m,sharma2018cc3m}, the Cauldron~\cite{laurencon2024idefics2}, LLaVA-OneVision~\cite{li2024llavaonevision}, Docmatix~\cite{laurencon2024building}, Aguvis~\cite{xu2024aguvis}, ChartGemma, ChartQA, ArxivQA and MathV360K~\cite{masry2024chartgemma,masry2022chartqa,li2024arxivqa,shi2024mathllava}, sampled roughly in proportion to their size (\suppTraining). As sampled, an image carries 254 image tokens and 86 text tokens on average, and OCR images account for 28\% of the image-text tokens. The text half draws from FineWeb-Edu~\cite{penedo2024fineweb}, OpenWebMath~\cite{paster2023openwebmath}, StarCoder data~\cite{li2023starcoder} and English Wikipedia.

\paragraph{Optimization.}
We train with AdamW~\cite{loshchilov2019adamw} with $\beta=(0.9,0.95)$ and weight decay 0.1, 2{,}000 warmup steps and cosine decay to 10\% of the peak learning rate over the token budget of the run (peak learning rates in \suppTraining). The budget is 40B tokens for the ladder and 20B for the interventions of \cref{sec:conditions}, and we call the two the 40B and the 20B schedule. At the same number of tokens the two schedules are at different learning rates, so losses are compared only between runs on the same schedule, and VIG comparisons that cross schedules are marked as such.

\subsection{Conditions}
\label{sec:conditions}

\paragraph{Frozen encoder.} In the frozen-encoder condition the linear embedding is replaced by a frozen SigLIP~2 so400m encoder~\cite{tschannen2025siglip2} at native aspect ratio, a $2\times2$ convolutional pooling and an MLP projector. Decoder, data and batch are unchanged. Chen \etal~\cite{chen2026howfar} train a SigLIP~2 encoder of this size jointly with the decoder. We keep it frozen, so the condition measures what the pretrained weights supply without any adaptation to the decoder. We run it at 323M on the 40B schedule of the ladder and at 158M on the 20B schedule of the interventions. The shuffle and the uniform image are applied to the pixels, before the encoder.

\paragraph{Interventions on the encoder-free decoder at 158M.}
\emph{Caption only} restricts the image-text half to CC12M and CC3M. \emph{Document-heavy} reweights the image-text half toward Docmatix, LLaVA-OneVision, the Cauldron and Aguvis, which doubles the text tokens per image and the OCR supervision per training token. \emph{16-pixel patches} quadruples the number of image tokens per image at the same token budget. \emph{Random-init ViT} inserts a trainable SigLIP~2 so400m architecture with random weights in place of the linear embedding, which separates the architecture from the pretrained weights. \emph{Batch size} trains with 512 sequences per step on the 40B schedule, at the peak learning rate and with the warmup tokens of the baseline. Batch 32 is in \suppSchedule. Two seeds of the 158M decoder with the default mixture on the 20B schedule are the baseline for these runs and give the run-to-run variation. The 158M decoder of the ladder, on the 40B schedule, is called the ladder baseline.

\section{What encoder-free pretraining learns}
\label{sec:results}

\begin{figure*}[t]
  \centering
  \includegraphics[width=\textwidth]{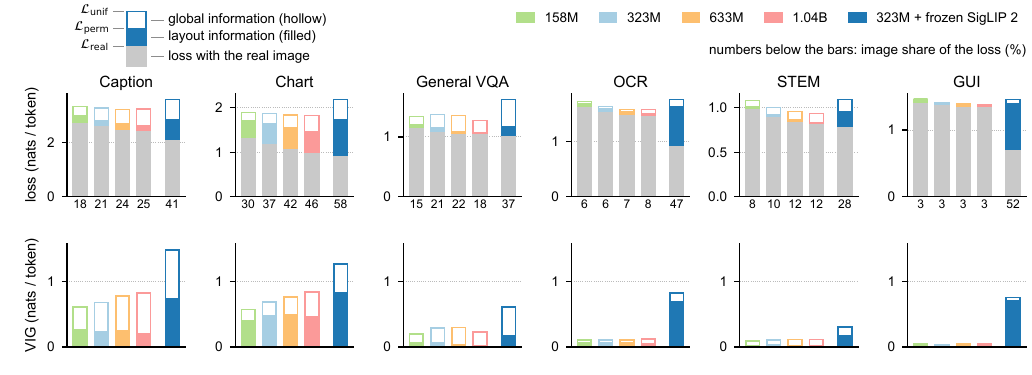}
  \caption{\textbf{Decomposition of the multimodal loss at 40B tokens.} Top: the loss with the real image (gray) with layout (filled) and global information (hollow) stacked on it, and below each bar the image share of the loss in percent. Bottom: the two parts alone, on a common scale. Each panel shows the four encoder-free sizes and the 323M decoder with the frozen SigLIP~2 encoder.}
  \label{fig:share}
\end{figure*}

\subsection{The encoder-free loss is mostly language}
\label{sec:language}

How much of the multimodal loss depends on the image? With a uniform image the decoders reach the image-blind loss $\Lblank$ from the text alone. The real image lowers it by $\vigtot$, and we call the relative reduction, $\vigtot/\Lblank$, the image share of the loss. After 40B tokens it ranges from 3\% on GUI screenshots and 6\% to 8\% on OCR images to 30\% to 46\% on charts across the four model sizes (\cref{fig:share}, top). The small share is not a property of the data, since with the frozen encoder it is 47\% on OCR images and 52\% on GUI screenshots.

A falling loss is therefore weak evidence of visual learning. On OCR images the loss of the encoder-free decoder falls by 0.83 nats between 1B and 40B tokens, while replacing the image by a uniform one never raises it by more than 0.11 nats, so almost the whole descent is language (\cref{fig:teaser}). The encoder does not change the language side either. On text-only sequences the two front ends reach the same loss at every checkpoint (1.975 against 1.977 nats at 40B tokens, 323M), as do the text frontiers of the two families in Chen \etal~\cite{chen2026howfar}, so they differ only in their use of the image.

A larger decoder does not help where the image share is smallest. From 158M to 1.04B the share rises from 30\% to 46\% on charts and from 18\% to 25\% on captions, and it stays flat on GUI screenshots and OCR images. \Cref{sec:clocks} separates the effect of model size from that of batch size.

\subsection{Two clocks for global and layout information}
\label{sec:clocks}

\begin{figure}[t]
  \centering
  \includegraphics[width=\linewidth]{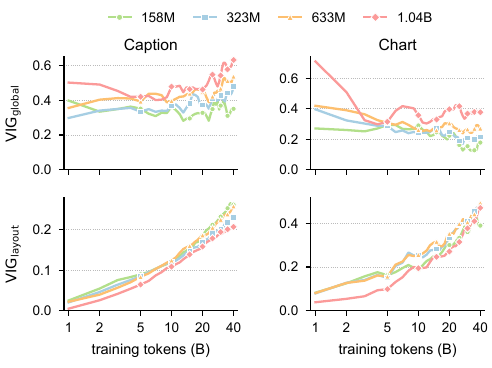}
  \caption{\textbf{Global information is present from the start, layout information grows with training.} $\vigglob$ (top) and $\viglay$ (bottom) on captions and charts against training tokens for the four encoder-free sizes, smoothed over three checkpoints.}
  \label{fig:clocks}
\end{figure}

When does a decoder start to use the image, and does a larger one start sooner? \Cref{fig:clocks} follows the two parts over training. Global information comes early. It is present after one billion tokens, at 0.30 to 0.50 nats on captions across the four sizes, and on charts it falls over training while layout information rises. At 40B tokens it is larger for the larger models on captions and charts, 0.62 nats at 1.04B against 0.34 at 158M on captions (\cref{fig:share}, bottom). Most of this difference comes with the batch size, which grows along the ladder. At 158M, batch 512 instead of 128 raises $\vigglob$ on captions from 0.34 to 0.52 nats. At equal batch size the larger model is ahead by 0.06 to 0.12 nats on captions and charts, the size of the seed-to-seed variation (\suppSchedule).

Layout information comes late. After one billion tokens it is near zero, at most 0.03 nats on captions and below 0.005 on OCR images, GUI screenshots and STEM figures. It then grows steadily, without a transition, and on most types is still rising when training ends. A larger model does not acquire it faster. At 40B tokens $\viglay$ lies between 0.21 and 0.27 nats on captions and between 0.04 and 0.07 on OCR images and does not increase with model size (\cref{fig:share}, bottom). At equal batch size the sizes differ by at most 0.04 nats on five types. On charts the 1.04B decoder ends 0.11 nats above the 158M one, about twice the difference between two seeds (\suppSchedule). Layout information is acquired with training, image type by image type. On OCR images and GUI screenshots it is still an order of magnitude below what a frozen encoder supplies (\cref{sec:encoder}), so every encoder-free run is in the bootstrap phase at 40B tokens.

\paragraph{Same loss, different use of the image.}
Can two runs with the same loss differ in how they use the image? We train the 158M decoder with batch 512 instead of 128, four times fewer optimizer steps for the same tokens. Below a critical batch size the loss depends on the tokens seen and not on the batch~\cite{mccandlish2018empirical,zhang2025critical}. The multimodal validation loss is unchanged (1.433 against 1.432 nats at 40B tokens), and the loss with the real image agrees within 0.03 nats on every image type (\suppSchedule). The loss with the shuffled image does not agree. It is lower with batch 512 on every type, by 0.02 to 0.06 nats, so this run has less layout information, 0.20 against 0.27 nats on captions and 0.04 against 0.07 on OCR images, and more global information. It relies more on the statistics of the patches and less on their arrangement. The layout effect exceeds the variation between two seeds on captions, OCR images and GUI screenshots, and the global effect on captions. The difference also shows in the answers. The run with batch 512 gets the leading digits of 39.5\% of the GUI coordinates right, against 44.6\% (\suppAnswers).

\subsection{What a pretrained encoder buys}
\label{sec:encoder}

\begin{figure*}[t]
  \centering
  \includegraphics[width=\textwidth]{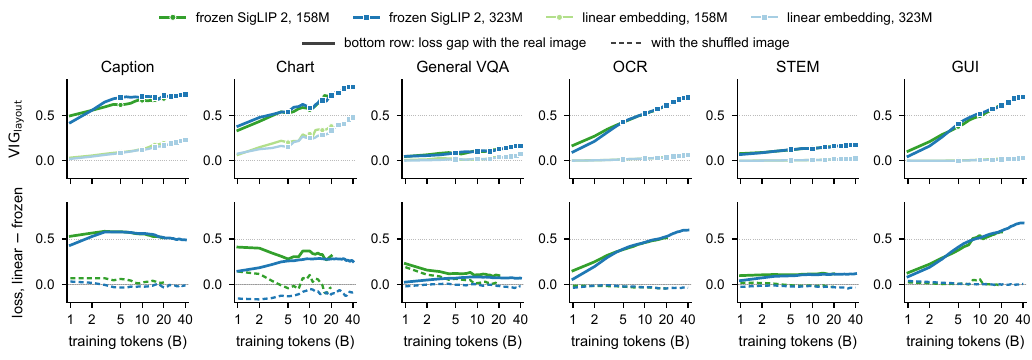}
  \caption{\textbf{The encoder contribution is layout information.} Top: $\viglay$ for the frozen-encoder condition (saturated) and the linear embedding (pastel) at 158M and 323M. Bottom: the loss of the linear embedding minus that of the frozen-encoder condition, with the real image (solid) and the shuffled image (dashed), which is near zero from 10B tokens on, except on charts. Curves are smoothed over three checkpoints. The 158M runs are on the 20B schedule, the 323M runs on the 40B schedule.}
  \label{fig:encoder}
\end{figure*}

\begin{table}[t]
  \caption{\textbf{The loss under the three image conditions with and without a pretrained encoder.} 323M decoder with the linear embedding or the frozen SigLIP~2 encoder at 40B tokens. With the shuffled image the two front ends agree on five of six types. Nats per text token, means over the last three checkpoints.}
  \label{tab:encoder}
  \centering
  \footnotesize
  \setlength{\tabcolsep}{2pt}
  \begin{tabular}{@{}lcccccc@{}}
\toprule
 & Caption & Chart & General & OCR & STEM & GUI \\
\midrule
\multicolumn{7}{@{}l}{\textit{Loss}} \\
Uniform image, linear & 3.29 & 1.86 & 1.37 & 1.62 & 1.00 & 1.42 \\
Uniform image, frozen & 3.61 & 2.18 & 1.62 & 1.75 & 1.09 & 1.46 \\
Shuffled image, linear & 2.84 & 1.65 & 1.16 & 1.58 & 0.92 & 1.41 \\
Shuffled image, frozen & 2.85 & 1.74 & 1.18 & 1.62 & 0.96 & 1.40 \\
Real image, linear & 2.61 & 1.17 & 1.09 & 1.52 & 0.90 & 1.38 \\
Real image, frozen & 2.12 & 0.91 & 1.02 & 0.92 & 0.78 & 0.70 \\
\addlinespace[2pt]
\multicolumn{7}{@{}l}{\textit{Information}} \\
Global, linear & 0.453 & 0.218 & 0.216 & 0.040 & 0.077 & 0.008 \\
Global, frozen & 0.756 & 0.443 & 0.445 & 0.130 & 0.133 & 0.054 \\
Layout, linear & 0.229 & 0.475 & 0.069 & 0.064 & 0.025 & 0.030 \\
Layout, frozen & 0.733 & 0.828 & 0.162 & 0.698 & 0.176 & 0.702 \\
\bottomrule
\end{tabular}

\end{table}

A frozen SigLIP~2 encoder lowers the multimodal validation loss of the 323M decoder by a quarter relative to the linear embedding, from 1.349 to 1.005 nats at 40B tokens. What does it supply? The shuffle keeps every patch and destroys their arrangement. If the encoder helped with what the patches contain, its lower loss would survive the shuffle. If it helped with how they are arranged, the two front ends would reach the same loss with the shuffled image.

They do. With the shuffled image the losses agree to within 0.04 nats on five of six types, and the frozen-encoder condition is 0.09 nats worse on charts (\cref{tab:encoder}). From 10B tokens on these five types stay within 0.05 nats, as do all six for the 158M pair at 20B tokens (\cref{fig:encoder}, bottom). On OCR images and GUI screenshots the agreement holds document by document: the frozen-encoder condition has the lower loss on 44\% and 47\% of them, about half (\suppEstimator).

What the encoder buys is therefore layout information. Since $\Lreal=\Lperm-\viglay$, equal losses with the shuffled image mean that its advantage in the loss is its lead in $\viglay$. On OCR images the lead is 0.63 nats and the advantage 0.60, and on GUI screenshots and captions the two agree within 0.01. The lead is there after one billion tokens, at 0.42 against 0.02 nats on captions. At 40B tokens the frozen-encoder condition has 0.70 nats of layout information on OCR images and GUI screenshots, over ten times what the linear embedding has (\cref{fig:encoder}, top, and \cref{tab:encoder}).

\paragraph{The frozen-encoder condition relies more on the image.}
The global information of the frozen-encoder condition is higher, by 0.3 nats on captions (\cref{tab:encoder}), but so is its image-blind loss. A decoder that has learned to rely on the image predicts worse without it, and the two effects cancel. $\vigtot$ therefore mixes what a model extracts from the image with how much it has come to depend on it, and should not be compared between two models as it stands. The layout gap is free of this, since it equals the gap in the loss with the real image. Even with the frozen encoder, $\viglay$ on OCR images rises only gradually, from 0.10 to 0.70 over training, so the decoder and the projector need data to use features that are there from the first step. The ability to read text emerges late in contrastive encoder training~\cite{park2025emergence}, and the frozen encoder brings it along, paid for in its own pretraining.

\paragraph{Layout information shows in the answers.}
VIG is a difference of losses, so we check it against the answers. We take the answer tokens that have to be read off the image, the digits in the answers on OCR images and the leading digits of GUI coordinates, which place the target in a cell of a $10\times10$ grid, and count how often the model ranks the correct token first under teacher forcing (\suppAnswers). On the OCR digits the image is worth 27.5 points to the frozen-encoder condition (89.9\% with the real image against 62.4\% with the uniform one) and 1.7 to 3.4 points to the encoder-free decoders across the four sizes. On the GUI coordinates, it is worth 46.7 points against 7.4 to 10.8. With the shuffled image the gains of the encoder-free decoders disappear. Accuracy thus separates the two front ends as $\viglay$ does.

\subsection{What does not replace the encoder}
\label{sec:interventions}

\begin{table}[t]
  \caption{\textbf{Interventions at 158M.} $\viglay$ in nats per text token at 20B tokens on the 20B schedule (top) and at 8B tokens (bottom). The 16-pixel run is shuffled in its own 16-pixel patches.}
  \label{tab:interventions}
  \centering
  \footnotesize
  \setlength{\tabcolsep}{1.6pt}
  \begin{tabular}{@{}lcccccc@{}}
\toprule
 & Caption & Chart & General & OCR & STEM & GUI \\
\midrule
\multicolumn{7}{@{}l}{\textit{20B tokens, 20B schedule}} \\
Baseline, seed 0 & 0.193 & 0.359 & 0.036 & 0.041 & 0.014 & 0.024 \\
Baseline, seed 1 & 0.166 & 0.312 & 0.031 & 0.039 & 0.011 & 0.016 \\
Caption only & 0.325 & 0.023 & 0.013 & $-$0.002 & 0.001 & $-$0.002 \\
Document-heavy & 0.090 & 0.339 & 0.014 & 0.043 & 0.011 & 0.017 \\
16-px patches & 0.103 & 0.239 & 0.067 & 0.014 & 0.008 & 0.004 \\
Frozen SigLIP~2 & 0.690 & 0.702 & 0.130 & 0.597 & 0.156 & 0.602 \\
\addlinespace[2pt]
\multicolumn{7}{@{}l}{\textit{8B tokens}} \\
Ladder baseline & 0.104 & 0.195 & 0.006 & 0.018 & 0.004 & 0.005 \\
Random-init ViT & 0.096 & 0.099 & 0.019 & 0.007 & 0.005 & 0.001 \\
\bottomrule
\end{tabular}

\end{table}

Can something cheaper than a pretrained encoder supply layout information? At 158M we test more supervision, finer patches and the encoder architecture without its weights (\cref{tab:interventions}). None does.

\paragraph{More OCR supervision does not improve reading.}
The document-heavy mixture carries, per training token, 1.9 times the OCR text and 1.6 times the OCR images of the default mixture. It draws on the same documents and shows them about twice as often, so it tests more supervision per document and not more documents. After 20B tokens $\viglay$ on OCR images is 0.04 nats, as for both baseline seeds. On captions more supervision does help. The caption-only mixture gives captions about twice the tokens, and their $\viglay$ reaches 0.33 against 0.17 and 0.19 for the seeds. Supervision thus helps where the decoder already extracts layout information, and not on OCR images.

\paragraph{Finer patches do not improve reading.}
With $16\times16$ patches the decoder sees each image with four times as many tokens and, at the same token budget, about three times fewer images. $\viglay$ on OCR images is 0.01 nats after 20B tokens, below the 0.04 of the baseline seeds. The comparison holds the training tokens fixed and not the number of images, so finer patches do not pay at equal compute.

\paragraph{The encoder architecture without its weights does not help within 8B tokens.}
A trainable SigLIP~2 so400m architecture with random weights adds 413M parameters in front of the decoder and triples the compute of the run. Through 8B tokens it extracts no more image information than the linear embedding, with $\viglay$ of 0.007 against 0.018 on OCR images and $\vigglob$ of 0.21 against 0.31 on captions, while the frozen-encoder condition reaches 0.47 on OCR images at the same point. Over this stretch, what the frozen encoder contributes comes with its pretrained weights and not with its architecture (\cref{sec:scaling}).

\section{Layout information in public models}
\label{sec:public}

\begin{figure}[t]
  \centering
  \includegraphics[width=\linewidth]{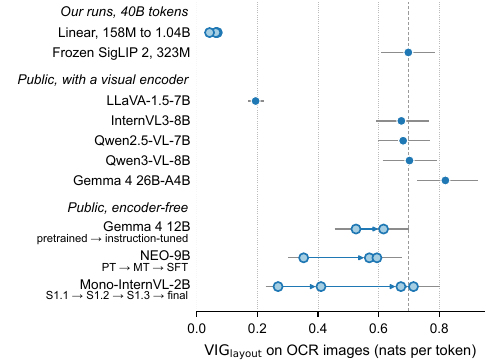}
  \caption{\textbf{Layout information on OCR images in our runs and in public checkpoints}, on the same documents and text. Filled markers are models with a visual encoder, light markers encoder-free models. Arrows connect released training stages. For Gemma~4 12B the arrow leads from the pretrained release, scored in the raw format, to the instruction-tuned release, scored in the chat format. The dashed line marks our frozen-encoder condition. Whiskers are bootstrap 95\% intervals over documents.}
  \label{fig:public}
\end{figure}

VIG needs only losses, so it applies to released checkpoints. We score the validation samples with 19 public checkpoints, each with its own image preprocessing. The summed loss of each model on the scored text is divided by the number of tokens of that text in our own evaluation (\suppPublic). Losses of different models are not comparable, since tokenizers and training text differ. Loss differences on the same text are comparable, provided that the scored text lies in the training distribution of the model. This holds on OCR images, where the image-blind loss of 18 of the 19 checkpoints lies within about 0.3 nats of the range of our decoders, so \cref{fig:public} shows OCR images for a selection of the checkpoints (all of them in \suppPublic).

Recent models with a visual encoder, InternVL3~\cite{zhu2025internvl3}, Qwen2.5-VL~\cite{bai2025qwen25vl}, Qwen3-VL~\cite{bai2025qwen3vl} and Gemma~4 26B-A4B~\cite{gemma2026gemma4}, have 0.65 to 0.82 nats of layout information. LLaVA-1.5~\cite{liu2024llava15}, whose CLIP encoder sees the image at 336 pixels, has 0.19. Our frozen-encoder condition lies inside the range of the recent models, at 0.70 with a 323M decoder after 40B tokens, and our encoder-free decoders lie an order of magnitude below it. Public encoder-free models do not stay there. Gemma~4 12B has 0.53 to 0.76 nats, depending on release, format and image-token budget, and the final releases of NEO~\cite{diao2025neo} and Mono-InternVL~\cite{luo2025monointernvl} have 0.60 to 0.72. The two ranges overlap, with the encoder-free models toward the lower end. Within the Gemma~4 family, at the same budget of about 260 image tokens, the encoder-based 26B-A4B is 0.20 nats above the encoder-free 12B (\suppPublic), and the encoder-free 12B scores below the smaller encoder-based 4.5B model of the family on document parsing at a matched budget of 280 visual tokens~\cite{gemma2026gemma4}.

The released training stages show layout information being acquired. It rises from 0.27 nats after the first stage of Mono-InternVL-2B to 0.72 in the final model, and from 0.35 to 0.60 over the three stages of NEO-9B. The first released stages of these two models already have four to eight times the layout information of our decoders at 40B tokens, so our runs lie below the earliest released stage, and the final releases have left the bootstrap phase. The stages differ in data and in which weights are trained as well as in compute, so they show that the gap closes and not which ingredient closes it. Model size alone is an unlikely candidate. In our ladder layout information on OCR images does not grow with model size (\cref{sec:clocks}), and the 2B models reach the level of the 9B and 12B ones. NEO and Mono-InternVL start from a pretrained language model and train in stages on several hundred million image-text examples or more, against the 19.8M samples of our mixture, and several encoder-free models align their patch features to a pretrained encoder during training~\cite{diao2024eve,tao2025hovle,wang2025vora}.

\section{Implications for scaling comparisons}
\label{sec:scaling}

Shukor \etal~\cite{shukor2025scaling} find no inherent advantage of a separate encoder when both families are trained from scratch, and the random-init ViT of \cref{sec:interventions} is consistent with this at small scale. A gap measured against a pretrained encoder, as in Chen \etal~\cite{chen2026howfar}, then reflects the pretrained weights, and in our runs it is layout information.

\paragraph{Image types catch up in the order of the layout gap.}
Chen \etal predict that encoder-free models catch up first on STEM figures, then on charts, and considerably later on GUI, OCR and captions, and in their downstream evaluation the largest average gaps are on DocVQA, ChartQA and TextVQA. In our runs at 40B tokens and 323M parameters, the layout gap between the frozen encoder and the linear embedding is 0.15 nats on STEM figures, 0.35 on charts, 0.50 on captions, 0.63 on OCR images and 0.67 on GUI screenshots (general VQA is not among their five topics). The two types on which Chen \etal expect encoder-free models to catch up first have the two smallest gaps, and the three they place last have the three largest. The loss gap gives the same order, since it equals the layout gap (\cref{sec:encoder}), and VIG adds the content of the gap. Chen \etal attribute the order to how strongly each topic relies on pretrained visual representations. Our measurement agrees and names what is relied on, the arrangement of the image, with reading at the top.

\paragraph{How the gap closes.}
Chen \etal report a sharp drop of the multimodal loss that ends the slow early phase of their models. Our decoders, dense and trained with AdamW on public data, show no such drop in 40B tokens, and their $\viglay$ grows as a power of the training tokens (\suppGrowth). If this growth continued, reaching the layout information of the frozen encoder would take from about 130B tokens on charts to more than 900B on OCR images. Whether the gap closes by such steady growth or in a faster regime is open. The rise in attention to visual tokens that Chen \etal report across their loss drop suggests a transition, which VIG on checkpoints around the drop can test. \suppDigitize{} locates the drop in their loss curves and shows how the compute before it enters the fitted compute law.

\section{Conclusion}
\label{sec:conclusion}

The comparison of encoder-free and encoder-based pretraining on the next-token loss can be broken down by what the image contributes. Early in encoder-free training the loss is mostly a language loss. It can fall while the image contributes little, and it can agree between two runs that use the image differently. A pretrained encoder contributes layout information and with it the ability to read, and on shuffled images its loss advantage is gone. More OCR supervision, finer patches and the encoder architecture without its weights do not supply this prior in our runs. Public encoder-free models at production scale reach layout information close to that of encoder-based models, and which ingredient of their training closes the gap is the next question for the measurement. Visual information gain costs three forward passes per checkpoint and can accompany any scaling study of vision-language pretraining. The code is available at \url{https://github.com/gezhixuan/visual-information-gain}.

\section*{LLM Usage Statement}
Claude (Anthropic) was used as an assistant for code, data
preparation, the analysis of results, figure scripts, literature
search, discussion of ideas and the text, in each case to edit or
refine material produced by the authors. Claude and ChatGPT (OpenAI)
gave reviewer-style feedback on drafts. The vision-language models
trained and evaluated in our experiments are objects of study and are
described in the main text. The authors verified all results and take
full responsibility for the content.

{
    \small
    \bibliographystyle{ieeenat_fullname}
    \bibliography{main}
}

\clearpage
\appendix
\maketitlesupplementary
\setcounter{section}{0}\renewcommand{\thesection}{\Alph{section}}\renewcommand{\theHsection}{supp.\Alph{section}}

\section{Estimator details}
\label{supp:estimator}

\paragraph{Validation batches.}
For each image type we pack 32 sequences of 4{,}096 tokens from held-out shards of the training sources with the same packing, loss mask and document attention as in training, and keep the packed batches fixed for every checkpoint and run (\cref{tab:supp_val}). The shuffle of each image is drawn once with a fixed seed and reused for every checkpoint and run, so the encoder-free decoders and the frozen-encoder condition are scored on the same shuffled images. The 16-pixel run is scored on the same samples. Every image is cut into 16-pixel patches, the samples are repacked into sequences of 8{,}192 tokens, and the shuffled image permutes the 16-pixel patches.

\begin{table}[h]
  \caption{\textbf{Validation batches per image type.} 32 packed sequences of 4{,}096 tokens each. Tokens are the scored text tokens on which the loss is computed. They include the end-of-image and the end-of-sample token of each sample. The chart and STEM types also hold the chart and STEM samples of the Cauldron and LLaVA-OneVision (72 of the 286 and 63 of the 204 images).}
  \label{tab:supp_val}
  \centering
  \footnotesize
  \setlength{\tabcolsep}{3.5pt}
  \begin{tabular}{@{}llrr@{}}
    \toprule
    Image type & Sources & Images & Tokens \\
    \midrule
    Caption & CC12M, CC3M & 609 & 11{,}689 \\
    Chart & ChartGemma, ChartQA & 286 & 12{,}209 \\
    General VQA & LLaVA-OV, Cauldron & 372 & 61{,}559 \\
    OCR & Docmatix & 96 & 33{,}003 \\
    STEM & ArxivQA, MathV360K & 204 & 44{,}668 \\
    GUI & Aguvis & 128 & 27{,}042 \\
    \bottomrule
  \end{tabular}
\end{table}

\paragraph{The noise condition.}
With pixels drawn uniformly at random in place of the uniform image, the loss of the four encoder-free decoders at 40B tokens differs from $\Lblank$ by at most 0.08 nats on captions and by at most 0.04 on the other types, in either direction. In the frozen-encoder condition of the 323M decoder, random pixels raise the loss above $\Lblank$ by 0.08 to 0.13 nats on captions, charts, OCR images and GUI screenshots, so the uniform image is the milder of the two references and we use it throughout.

\paragraph{A mismatched image as reference.}
The uniform image and the shuffled image are inputs that a model has not seen in training. As a further reference we replace the image of every sample by the image of another sample of the same type, resized to the size of the original, so that the number of image tokens is unchanged. The samples of a type are put in a fixed random order, each takes the image of the next one, and a partner with an identical image is skipped. \Cref{tab:supp_mismatch} gives the loss increase over the real image at the 40B checkpoint. On OCR images it is 0.08 to 0.09 nats for the encoder-free decoders and 0.89 for the frozen-encoder condition, against 0.10 to 0.11 and 0.82 with the uniform image. On GUI screenshots it is 0.05 to 0.06 and 0.81, against 0.04 to 0.05 and 0.76. The frozen-encoder condition lies above the encoder-free decoders on every image type under either reference, by an order of magnitude on OCR images and GUI screenshots. For the frozen-encoder condition the mismatched image is the harsher reference on five types, most of all on captions (2.09 against 1.49 nats), since an image with other content misleads a model that uses it. For the encoder-free decoders on charts it is the milder one, 0.36 to 0.59 nats against 0.56 to 0.86, so part of what these decoders take from a chart is shared by the charts of the validation set and is not specific to the sample. The order of the two front ends does not depend on the checkpoint. For the 323M pair at 1, 5, 10 and 20B tokens the mismatched image costs the encoder-free decoder at most 0.06 nats on OCR images and the frozen-encoder condition 0.20 to 0.80. The encoder-free Gemma~4 12B and the encoder-based Qwen3-VL-8B (Appendix~\ref{supp:public}), scored on OCR images and GUI screenshots, behave like the frozen-encoder condition. On OCR images the mismatched image costs them 0.70 and 0.82 nats, against 0.69 and 0.79 with the uniform image, and on GUI screenshots it is the harsher reference by 0.02 and 0.07 nats.

\begin{table*}[t]
  \caption{\textbf{A mismatched image as reference.} Loss increase over the real image when the image is replaced by the image of another sample of the same type (Mis.) and by the uniform image (Unif., which is $\vigtot$), in nats per scored token of our evaluation. Our decoders at the 40B checkpoint and two public models, which are scored on OCR images and GUI screenshots.}
  \label{tab:supp_mismatch}
  \centering
  \small
  \setlength{\tabcolsep}{4pt}
  \begin{tabular}{@{}lcccccccccccc@{}}
\toprule
 & \multicolumn{2}{c}{Caption} & \multicolumn{2}{c}{Chart} & \multicolumn{2}{c}{General} & \multicolumn{2}{c}{OCR} & \multicolumn{2}{c}{STEM} & \multicolumn{2}{c}{GUI} \\
\cmidrule(lr){2-3}\cmidrule(lr){4-5}\cmidrule(lr){6-7}\cmidrule(lr){8-9}\cmidrule(lr){10-11}\cmidrule(l){12-13}
Reference image & Mis. & Unif. & Mis. & Unif. & Mis. & Unif. & Mis. & Unif. & Mis. & Unif. & Mis. & Unif. \\
\midrule
158M, linear & 0.743 & 0.596 & 0.357 & 0.559 & 0.179 & 0.197 & 0.075 & 0.107 & 0.102 & 0.083 & 0.058 & 0.047 \\
323M, linear & 0.801 & 0.717 & 0.467 & 0.694 & 0.290 & 0.288 & 0.085 & 0.101 & 0.119 & 0.100 & 0.053 & 0.041 \\
633M, linear & 0.864 & 0.800 & 0.561 & 0.767 & 0.251 & 0.300 & 0.091 & 0.109 & 0.121 & 0.113 & 0.054 & 0.046 \\
1.04B, linear & 0.845 & 0.857 & 0.594 & 0.864 & 0.179 & 0.222 & 0.085 & 0.114 & 0.121 & 0.112 & 0.050 & 0.042 \\
323M, frozen SigLIP~2 & 2.094 & 1.485 & 0.997 & 1.279 & 0.673 & 0.614 & 0.888 & 0.821 & 0.366 & 0.314 & 0.808 & 0.756 \\
\addlinespace[2pt]
Gemma~4 12B, raw, 1{,}082 tokens & -- & -- & -- & -- & -- & -- & 0.702 & 0.690 & -- & -- & 0.635 & 0.615 \\
Qwen3-VL-8B, chat & -- & -- & -- & -- & -- & -- & 0.816 & 0.792 & -- & -- & 0.809 & 0.738 \\
\bottomrule
\end{tabular}

\end{table*}

\paragraph{Block size of the shuffle.}
The shuffle of the main evaluation moves blocks of 32 pixels, the patch size of the linear embedding. \Cref{tab:supp_scale} repeats the evaluation of the 40B checkpoints with blocks of 8 to 128 pixels and new random permutations. At 32 pixels the new permutations reproduce the values of the main evaluation to within 0.015 nats on every image type. Blocks of 64 and 128 pixels keep more of the arrangement and the loss increase shrinks, on OCR images from 0.70 to 0.42 nats for the frozen-encoder condition and from 0.07 to 0.05 for the 323M encoder-free decoder, so the two differ by an order of magnitude at every block size. Blocks smaller than the patch change the content of every patch. On GUI screenshots the loss increase of the encoder-free decoders then exceeds that of the uniform image, 0.09 against 0.04 nats for the 323M decoder, and on OCR images it comes within 0.03 nats of it, so below the patch size the shuffle no longer separates layout from global information. Gemma~4 12B and Qwen3-VL-8B (Appendix~\ref{supp:public}) follow the profile of the frozen-encoder condition, with 0.62 and 0.70 nats at 32 pixels and 0.28 and 0.31 at 128 on OCR images.

\begin{table*}[t]
  \caption{\textbf{Block size of the shuffle.} Loss increase of the image shuffled in blocks of the given size over the real image, and $\vigtot$ for reference, in nats per scored token of our evaluation. Our decoders at the 40B checkpoint and two public models.}
  \label{tab:supp_scale}
  \centering
  \small
  \setlength{\tabcolsep}{4pt}
  \begin{tabular}{@{}lcccccccccccc@{}}
\toprule
 & \multicolumn{6}{c}{OCR} & \multicolumn{6}{c}{GUI} \\
\cmidrule(lr){2-7}\cmidrule(l){8-13}
Block size (pixels) & 8 & 16 & 32 & 64 & 128 & $\vigtot$ & 8 & 16 & 32 & 64 & 128 & $\vigtot$ \\
\midrule
323M, linear & 0.095 & 0.095 & 0.067 & 0.057 & 0.051 & 0.101 & 0.087 & 0.080 & 0.031 & 0.026 & 0.025 & 0.041 \\
1.04B, linear & 0.099 & 0.086 & 0.043 & 0.036 & 0.032 & 0.114 & 0.077 & 0.073 & 0.021 & 0.018 & 0.015 & 0.042 \\
323M, frozen SigLIP~2 & 0.825 & 0.769 & 0.705 & 0.589 & 0.422 & 0.821 & 0.815 & 0.757 & 0.710 & 0.634 & 0.505 & 0.756 \\
\addlinespace[2pt]
Gemma~4 12B, raw, 1{,}082 tokens & 0.713 & 0.695 & 0.622 & 0.477 & 0.279 & 0.691 & 0.649 & 0.597 & 0.483 & 0.307 & 0.164 & 0.615 \\
Qwen3-VL-8B, chat & 0.778 & 0.780 & 0.702 & 0.518 & 0.309 & 0.792 & 0.729 & 0.736 & 0.660 & 0.503 & 0.353 & 0.738 \\
\bottomrule
\end{tabular}

\end{table*}

\paragraph{Checkpoint-to-checkpoint noise.}
\Cref{tab:supp_noise} gives the standard deviation of consecutive differences divided by $\sqrt{2}$ over the last 16 checkpoints of the 158M and 323M runs. $\viglay$ is quiet on every type except charts. $\vigglob$ carries the noise of the image-blind loss on captions and charts, which have the fewest scored tokens. Values quoted at the end of a run in the main text are means over its last three checkpoints, and the curves in \cref{fig:clocks,fig:encoder} are smoothed over three checkpoints, with the first checkpoint left as measured. Two evaluations of the same checkpoint on two machines agree to 0.0003 nats.

\begin{table}[h]
  \caption{\textbf{Checkpoint noise} in nats per text token for the 158M / 323M encoder-free decoders.}
  \label{tab:supp_noise}
  \centering
  \footnotesize
  \setlength{\tabcolsep}{3.5pt}
  \begin{tabular}{@{}lccc@{}}
    \toprule
    Image type & $\viglay$ & $\vigglob$ & $\vigtot$ \\
    \midrule
    Caption & 0.004 / 0.004 & 0.052 / 0.028 & 0.053 / 0.028 \\
    Chart & 0.028 / 0.016 & 0.026 / 0.026 & 0.027 / 0.025 \\
    General VQA & 0.003 / 0.005 & 0.009 / 0.009 & 0.009 / 0.009 \\
    OCR & 0.003 / 0.001 & 0.006 / 0.005 & 0.006 / 0.005 \\
    STEM & 0.001 / 0.002 & 0.004 / 0.003 & 0.004 / 0.003 \\
    GUI & 0.002 / 0.001 & 0.003 / 0.004 & 0.003 / 0.003 \\
    \bottomrule
  \end{tabular}
\end{table}

\paragraph{Paired comparison over samples.}
Checkpoint noise and the two seeds describe the variation between checkpoints and between trainings. \Cref{tab:supp_paired} adds the uncertainty of the finite evaluation set. It uses the loss of every sample at a single checkpoint and a paired bootstrap over samples with 10{,}000 resamples, in which both models are scored on the same resampled images and text. With the real image the frozen-encoder condition has the lower loss on 99\% of the OCR documents and 98\% of the GUI screenshots. With the shuffled image it has the lower loss on 44\% and 47\%, and the losses of the two front ends differ by 0.09 and 0.05 nats per sample on average, against 0.62 and 0.61 with the real image and 0.04 between the two seeds of the 158M decoder. On these two types the shuffled-image losses therefore agree sample by sample. On captions they agree in the mean, with a difference of $-0.01$ nats and an interval from $-0.05$ to $+0.03$, while single captions differ by 0.38 nats on average, twice the difference between the two seeds. A caption has about 19 scored tokens. On charts, general questions, OCR images and STEM figures the frozen-encoder condition is 0.03 to 0.10 nats worse with the shuffled image and the interval excludes zero, against a lead in $\viglay$ of 0.10 to 0.64 nats on these types.

The lower block gives differences in $\viglay$. The lead of the frozen-encoder condition and the effect of the batch size exclude zero on all six types. Between model sizes at equal batch size the larger decoder is ahead by more than 0.03 nats only on charts, for the 1.04B decoder, and on general questions it is behind by 0.03 to 0.04. The values of this table are those of single checkpoints. On charts, where the checkpoint noise is largest, the two seeds differ by 0.09 nats at the 20B checkpoint and by 0.05 in the mean of the last three checkpoints, and the 1.04B decoder is 0.12 nats above the 158M decoder at the 40B checkpoint and 0.11 in the mean of the last three (Appendix~\ref{supp:schedule}). On captions and charts the two seeds differ by more than their intervals over samples, so the variation between trainings limits the small effects more than the size of the evaluation set does, and the main text measures them against the seeds.

\begin{table*}[t]
  \caption{\textbf{Paired comparison over samples.} Upper block: loss with the real and with the shuffled image of the 323M linear embedding minus that of the frozen-encoder condition at the 40B checkpoint, the share of samples on which the frozen-encoder condition has the lower loss, and the mean absolute difference per sample, in nats per scored token. Lower block: differences in $\viglay$ between two runs. Intervals are 95\% paired bootstrap intervals over samples.}
  \label{tab:supp_paired}
  \centering
  \footnotesize
  \setlength{\tabcolsep}{2.5pt}
  \begin{tabular}{@{}lcccccc@{}}
\toprule
 & Caption & Chart & General & OCR & STEM & GUI \\
\midrule
\multicolumn{7}{@{}l}{\textit{Loss of the linear embedding minus loss of the frozen-encoder condition, 323M, 40B tokens}} \\
Samples & 609 & 286 & 372 & 96 & 204 & 128 \\
Real image & $+0.488$ & $+0.246$ & $+0.069$ & $+0.598$ & $+0.118$ & $+0.675$ \\
\quad 95\% interval & [0.438, 0.543] & [0.215, 0.280] & [0.061, 0.077] & [0.526, 0.671] & [0.100, 0.138] & [0.594, 0.759] \\
\quad frozen-encoder loss lower on & 89\% & 85\% & 91\% & 99\% & 85\% & 98\% \\
Shuffled image & $-0.011$ & $-0.095$ & $-0.026$ & $-0.039$ & $-0.036$ & $+0.000$ \\
\quad 95\% interval & [$-$0.048, 0.026] & [$-$0.124, $-$0.065] & [$-$0.035, $-$0.017] & [$-$0.063, $-$0.015] & [$-$0.047, $-$0.025] & [$-$0.012, 0.012] \\
\quad frozen-encoder loss lower on & 50\% & 38\% & 40\% & 44\% & 36\% & 47\% \\
Mean absolute difference per sample & & & & & & \\
\quad real image & 0.59 & 0.32 & 0.13 & 0.62 & 0.15 & 0.61 \\
\quad shuffled image & 0.38 & 0.25 & 0.12 & 0.09 & 0.09 & 0.05 \\
\quad shuffled, two seeds (158M, 20B) & 0.18 & 0.21 & 0.09 & 0.04 & 0.07 & 0.04 \\
\addlinespace[3pt]
\multicolumn{7}{@{}l}{\textit{Difference in $\viglay$ and its 95\% interval}} \\
Frozen encoder $-$ linear, 323M, 40B & $+0.499$ & $+0.341$ & $+0.095$ & $+0.637$ & $+0.154$ & $+0.676$ \\
 & [0.446, 0.556] & [0.304, 0.380] & [0.085, 0.105] & [0.555, 0.719] & [0.134, 0.176] & [0.591, 0.762] \\
Batch 128 $-$ batch 512, 158M, 40B & $+0.067$ & $+0.048$ & $+0.017$ & $+0.029$ & $+0.012$ & $+0.029$ \\
 & [0.053, 0.082] & [0.031, 0.066] & [0.014, 0.021] & [0.023, 0.036] & [0.008, 0.016] & [0.025, 0.033] \\
633M $-$ 323M, batch 256, 40B & $+0.028$ & $+0.019$ & $-0.037$ & $-0.003$ & $-0.003$ & $+0.006$ \\
 & [0.015, 0.042] & [$-$0.003, 0.042] & [$-$0.042, $-$0.031] & [$-$0.010, 0.003] & [$-$0.008, 0.002] & [0.002, 0.009] \\
1.04B $-$ 158M, batch 512, 40B & $+0.012$ & $+0.117$ & $-0.027$ & $+0.003$ & $+0.004$ & $-0.007$ \\
 & [$-$0.002, 0.026] & [0.090, 0.148] & [$-$0.031, $-$0.023] & [$-$0.003, 0.009] & [0.001, 0.008] & [$-$0.010, $-$0.003] \\
Seed 0 $-$ seed 1, 158M, 20B & $+0.033$ & $+0.094$ & $-0.005$ & $+0.005$ & $+0.001$ & $+0.010$ \\
 & [0.022, 0.044] & [0.073, 0.116] & [$-$0.007, $-$0.002] & [0.001, 0.008] & [$-$0.002, 0.004] & [0.006, 0.014] \\
\bottomrule
\end{tabular}

\end{table*}

\section{Schedule-matched comparisons}
\label{supp:schedule}

The 158M baseline of the ladder uses a 40B-token cosine schedule. The interventions at 158M (caption only, document-heavy, 16-pixel patches, batch 32, the frozen-encoder condition and the two baseline seeds) use a 20B schedule, so at 20B tokens they are at a learning rate of $1.5\times10^{-4}$ while the ladder baseline is at $8.5\times10^{-4}$. The schedule alone changes the text validation loss at 20B tokens by about 0.1 nats (2.292 for the ladder baseline at 20B tokens against 2.179 and 2.181 for the two seeds at the end of the 20B schedule). \Cref{tab:supp_sched} therefore compares every 20B-schedule run with the baseline seeds on the same schedule, and lists the ladder baseline at 20B tokens for reference only. The frozen-encoder condition reaches a text validation loss of 2.167 and the document-heavy mixture 2.179, against 2.179 and 2.181 for the two seeds, so neither changes how well the decoder models text. The two baseline seeds differ in $\viglay$ by 0.02 to 0.03 nats on captions and 0.03 to 0.05 on charts at 10B and 20B tokens, and by at most 0.01 on the other four types. In $\vigglob$ they differ by 0.08 to 0.12 nats on captions, charts and general questions at 10B tokens and by up to 0.10 at 20B. $\vigglob$ therefore carries most of the run-to-run variation as well as most of the checkpoint noise.

\paragraph{Model size and batch size.}
The batch grows along the ladder (\cref{tab:ladder}), so model size is compared at equal batch size, between the 323M and 633M decoders and between the 158M decoder trained with batch 512 and the 1.04B decoder, at the end of training. In $\vigglob$ the larger model is ahead by 0.06 to 0.12 nats on captions and charts, which is the size of the difference between the two seeds. In $\viglay$ the sizes differ by at most 0.04 nats on five types. On charts the 1.04B decoder ends 0.11 nats above the 158M one, against 0.05 between the two seeds at the end of their run, and the difference opens late. Averaged over the second half of training it is 0.06. \Cref{tab:supp_batch} sets the 158M decoder trained with batch 512 beside the one trained with batch 128 at 40B tokens. Batch 512 raises $\vigglob$ by 0.18 nats on captions and 0.07 on charts at the end of training, and by 0.15 and 0.10 averaged over the second half. Its $\viglay$ is lower on every type, and the paired bootstrap over samples excludes zero on all six (\cref{tab:supp_paired}). On the leading digits of the GUI coordinates the run with batch 512 is right on 39.5\% against 44.6\%, and on the OCR digits, where both runs have little layout information, the two are within one point (\cref{tab:supp_answers}). The 1.04B decoder, trained with batch 512 and half the optimizer steps of the 323M and 633M decoders, has the lowest $\viglay$ at 40B tokens on five of the six types. $\viglay$ reaches half of its 40B value after 7 to 13B tokens on captions and charts and only after 20 to 23B on GUI screenshots, across the four sizes. In the multimodal validation loss the two seeds differ by 0.016 nats at 20B tokens, and the run with batch 32, four times the steps of the baseline, lies 0.09 to 0.11 nats above them. Its $\viglay$ stays within 0.02 nats of the range of the two seeds on every type, so below batch 128 extra steps add nothing.

\paragraph{The caption-only run.}
The caption-only run has not seen the text of the other image types. Its image-blind loss on them is 2.0 to 3.7 nats, 0.6 to 1.8 above that of the baseline, and its loss stays far above that of the baseline with any image. The run therefore shows what twice the caption supervision does for captions and does not test whether layout information carries over between image types.

\begin{table*}[t]
  \caption{\textbf{Schedule-matched comparisons at 158M and 20B tokens.} Means over the three checkpoints at or before 20B tokens, nats per text token.}
  \label{tab:supp_sched}
  \centering
  \small
  \begin{tabular}{@{}lcccccc@{}}
\toprule
 & Caption & Chart & General & OCR & STEM & GUI \\
\midrule
\multicolumn{7}{@{}l}{$\vigtot$} \\
Baseline, seed 0 & 0.560 & 0.565 & 0.210 & 0.084 & 0.082 & 0.033 \\
Baseline, seed 1 & 0.570 & 0.544 & 0.300 & 0.098 & 0.084 & 0.037 \\
Batch 32 & 0.505 & 0.487 & 0.338 & 0.065 & 0.077 & 0.036 \\
Caption only & 0.710 & $-$0.046 & $-$0.006 & 0.011 & $-$0.002 & 0.025 \\
Document-heavy & 0.482 & 0.596 & 0.171 & 0.105 & 0.078 & 0.030 \\
16-px patches & 0.506 & 0.464 & 0.436 & 0.062 & 0.081 & 0.037 \\
Frozen SigLIP~2 & 1.306 & 1.109 & 0.560 & 0.709 & 0.294 & 0.634 \\
Ladder baseline & 0.505 & 0.507 & 0.180 & 0.079 & 0.074 & 0.038 \\
\midrule
\multicolumn{7}{@{}l}{$\viglay$} \\
Baseline, seed 0 & 0.193 & 0.359 & 0.036 & 0.041 & 0.014 & 0.024 \\
Baseline, seed 1 & 0.166 & 0.312 & 0.031 & 0.039 & 0.011 & 0.016 \\
Batch 32 & 0.190 & 0.356 & 0.051 & 0.039 & 0.013 & 0.030 \\
Caption only & 0.325 & 0.023 & 0.013 & $-$0.002 & 0.001 & $-$0.002 \\
Document-heavy & 0.090 & 0.339 & 0.014 & 0.043 & 0.011 & 0.017 \\
16-px patches & 0.103 & 0.239 & 0.067 & 0.014 & 0.008 & 0.004 \\
Frozen SigLIP~2 & 0.690 & 0.702 & 0.130 & 0.597 & 0.156 & 0.602 \\
Ladder baseline & 0.180 & 0.308 & 0.032 & 0.037 & 0.012 & 0.018 \\
\midrule
\multicolumn{7}{@{}l}{$\Lblank$} \\
Baseline, seed 0 & 3.408 & 1.930 & 1.421 & 1.731 & 1.112 & 1.486 \\
Baseline, seed 1 & 3.434 & 1.954 & 1.511 & 1.753 & 1.114 & 1.476 \\
Batch 32 & 3.532 & 2.014 & 1.644 & 1.827 & 1.209 & 1.561 \\
Caption only & 3.302 & 3.683 & 2.695 & 2.352 & 2.021 & 2.467 \\
Document-heavy & 3.647 & 1.913 & 1.352 & 1.697 & 1.069 & 1.470 \\
16-px patches & 3.644 & 2.131 & 1.735 & 1.833 & 1.234 & 1.523 \\
Frozen SigLIP~2 & 3.641 & 2.174 & 1.671 & 1.845 & 1.210 & 1.512 \\
Ladder baseline & 3.491 & 1.975 & 1.435 & 1.810 & 1.188 & 1.514 \\
\bottomrule
\end{tabular}

\end{table*}

\begin{table*}[t]
  \caption{\textbf{Batch size at 158M on the 40B schedule.} The 158M decoder trained with batch 128 (the ladder baseline) and with batch 512 at 40B tokens. Means over the last three checkpoints, nats per text token. The last row of the two lower blocks is the absolute difference between the two baseline seeds at 20B tokens.}
  \label{tab:supp_batch}
  \centering
  \small
  \begin{tabular}{@{}lcccccc@{}}
\toprule
 & Caption & Chart & General & OCR & STEM & GUI \\
\midrule
\multicolumn{7}{@{}l}{\textit{Loss with the uniform image, $\Lblank$}} \\
Batch 128 & 3.358 & 1.890 & 1.348 & 1.717 & 1.078 & 1.454 \\
Batch 512 & 3.483 & 1.903 & 1.386 & 1.697 & 1.062 & 1.449 \\
\midrule
\multicolumn{7}{@{}l}{\textit{Loss with the shuffled image, $\Lperm$}} \\
Batch 128 & 3.014 & 1.707 & 1.211 & 1.678 & 1.014 & 1.464 \\
Batch 512 & 2.962 & 1.648 & 1.188 & 1.640 & 0.987 & 1.439 \\
\midrule
\multicolumn{7}{@{}l}{\textit{Loss with the real image, $\Lreal$}} \\
Batch 128 & 2.749 & 1.314 & 1.149 & 1.611 & 0.991 & 1.410 \\
Batch 512 & 2.767 & 1.290 & 1.143 & 1.601 & 0.976 & 1.412 \\
\midrule
\multicolumn{7}{@{}l}{\textit{Global information, $\vigglob$}} \\
Batch 128 & 0.344 & 0.183 & 0.136 & 0.038 & 0.064 & $-$0.010 \\
Batch 512 & 0.522 & 0.255 & 0.199 & 0.056 & 0.075 & 0.010 \\
Two seeds, difference & 0.037 & 0.026 & 0.096 & 0.016 & 0.004 & 0.012 \\
\midrule
\multicolumn{7}{@{}l}{\textit{Layout information, $\viglay$}} \\
Batch 128 & 0.265 & 0.393 & 0.062 & 0.068 & 0.023 & 0.054 \\
Batch 512 & 0.195 & 0.357 & 0.045 & 0.040 & 0.011 & 0.027 \\
Two seeds, difference & 0.027 & 0.047 & 0.006 & 0.002 & 0.002 & 0.008 \\
\bottomrule
\end{tabular}

\end{table*}

\section{Answer accuracy}
\label{supp:answers}

\paragraph{Protocol.}
VIG is a difference of losses. To check that layout information corresponds to reading, we score the answers directly. Except for captions, the text of every validation sample consists of questions and answers. For every answer token we record under teacher forcing whether the target is the top-ranked prediction, under the real, the shuffled and the uniform image, on the same batches and with the same shuffle as for VIG. An answer whose tokens are all ranked first is the answer that greedy decoding from the question produces, which we count as an exact match. \Cref{tab:supp_answers} reports four measures: the digits in the answers on OCR images (1{,}650 tokens), which have to be read off the document, the leading decimal digit of the $x$ and $y$ coordinates in the answers on GUI screenshots (1{,}378 tokens), which places the target in a cell of a $10\times10$ grid, and exact match on charts (296 answers) and on general questions (1{,}646 answers). The accuracy with the uniform image is what the text context alone gives.

\begin{table*}[t]
  \caption{\textbf{Answer accuracy under the three image conditions.} Share of answer tokens whose target is the top-ranked prediction under teacher forcing (digits, coordinates) and share of answers reproduced exactly (exact match), in percent.}
  \label{tab:supp_answers}
  \centering
  \small
  \setlength{\tabcolsep}{4pt}
  \begin{tabular}{@{}lcccccccccccc@{}}
\toprule
 & \multicolumn{3}{c}{OCR, digits} & \multicolumn{3}{c}{GUI, coordinates} & \multicolumn{3}{c}{Chart, exact match} & \multicolumn{3}{c}{General, exact match} \\
\cmidrule(l{3pt}r{3pt}){2-4}\cmidrule(l{3pt}r{3pt}){5-7}\cmidrule(l{3pt}r{3pt}){8-10}\cmidrule(l{3pt}r{3pt}){11-13}
 & uniform & shuffled & real & uniform & shuffled & real & uniform & shuffled & real & uniform & shuffled & real \\
\midrule
\multicolumn{13}{@{}l}{\textit{40B tokens}} \\
158M, linear & 66.4 & 66.8 & 68.6 & 34.0 & 30.4 & 44.6 & 3.0 & 5.1 & 6.4 & 38.9 & 44.7 & 52.7 \\
323M, linear & 66.2 & 66.4 & 69.6 & 31.9 & 31.2 & 42.7 & 2.0 & 3.7 & 8.4 & 40.8 & 46.4 & 57.5 \\
633M, linear & 68.5 & 68.6 & 70.7 & 33.4 & 30.7 & 41.2 & 2.4 & 5.7 & 10.8 & 43.6 & 51.5 & 56.9 \\
1.04B, linear & 70.0 & 70.8 & 71.7 & 34.0 & 33.8 & 41.4 & 2.7 & 8.1 & 11.1 & 45.7 & 52.3 & 54.7 \\
323M, frozen SigLIP~2 & 62.4 & 66.4 & 89.9 & 33.0 & 27.7 & 79.8 & 2.0 & 4.4 & 45.3 & 36.0 & 49.8 & 75.4 \\
158M, linear, batch 512 & 66.5 & 67.0 & 69.3 & 32.8 & 29.7 & 39.5 & 2.0 & 5.1 & 8.1 & 39.5 & 45.9 & 51.2 \\
\addlinespace[2pt]
\multicolumn{13}{@{}l}{\textit{158M, 20B tokens, 20B schedule}} \\
Baseline, seed 0 & 65.5 & 65.7 & 67.0 & 34.0 & 29.8 & 39.4 & 2.0 & 4.1 & 5.7 & 40.4 & 46.3 & 50.9 \\
Baseline, seed 1 & 65.4 & 65.0 & 66.2 & 32.9 & 33.1 & 38.6 & 2.0 & 4.4 & 8.1 & 40.2 & 45.3 & 49.2 \\
Document-heavy & 66.8 & 67.5 & 69.3 & 32.7 & 31.6 & 39.8 & 1.4 & 3.7 & 7.4 & 42.2 & 47.2 & 50.0 \\
16-px patches & 63.4 & 63.7 & 63.8 & 32.7 & 33.2 & 34.8 & 1.7 & 3.4 & 2.7 & 34.7 & 41.5 & 41.6 \\
Caption only & 59.5 & 59.6 & 59.2 & 21.6 & 21.6 & 21.8 & 0.0 & 0.0 & 0.0 & 3.9 & 3.6 & 3.8 \\
Frozen SigLIP~2 & 60.1 & 65.8 & 86.4 & 31.8 & 27.4 & 72.0 & 1.4 & 2.7 & 36.8 & 36.2 & 47.3 & 64.0 \\
\bottomrule
\end{tabular}

\end{table*}

\paragraph{Results.}
With the real image the frozen-encoder condition of the 323M decoder gains 27.5 points on the OCR digits (95\% bootstrap interval over samples 21.9 to 33.4), 46.7 on the GUI coordinates (42.9 to 50.4) and 43.2 on chart answers (37.5 to 49.0). The encoder-free decoders of the four sizes gain 1.7 to 3.4 points on the OCR digits, 7.4 to 10.8 on the GUI coordinates and 3.4 to 8.4 on chart answers, with no increase with model size on OCR images and GUI screenshots. Their intervals exclude zero, for the 323M decoder 1.4 to 5.8 on the digits and 8.8 to 12.9 on the coordinates, so the decoders have begun to use the layout of these images, at less than a quarter of the gain of the frozen-encoder condition. With the shuffled image the gains on OCR images and GUI screenshots disappear for the encoder-free decoders, and the frozen-encoder condition keeps 4 to 6 points on the OCR digits and none on the GUI coordinates. The 20B-schedule runs at 158M order as their layout information does. The document-heavy mixture gains 2.5 points on the OCR digits against 1.6 and 0.8 for the two baseline seeds, with overlapping intervals, the 16-pixel run gains 0.4, the caption-only run gains nothing on OCR images, GUI screenshots, charts and general questions, and the frozen-encoder condition gains 26.4 points. For the two batch sizes compared in the main text, at 158M and 40B tokens, the gain on the OCR digits is 2.2 points with batch 128 and 2.8 with batch 512, with overlapping intervals (0.5 to 3.9 and 1.3 to 4.5). On the GUI coordinates it is 10.6 against 6.7 points (8.1 to 13.3 and 4.3 to 9.0), with 44.6\% against 39.5\% of the leading digits correct with the real image. Over training (\cref{tab:supp_answers_traj}) the gain of the frozen-encoder condition on the OCR digits rises from 2 points after one billion tokens to 20 after 5B and 27 after 30B, and that of the encoder-free decoder stays below 4.

\begin{table}[h]
  \caption{\textbf{Answer accuracy over training.} Gain in accuracy of the real over the uniform image, in points, on the digits in OCR answers and on the leading digits of GUI coordinates, for the 323M decoder with the linear embedding and with the frozen encoder.}
  \label{tab:supp_answers_traj}
  \centering
  \footnotesize
  \setlength{\tabcolsep}{3.5pt}
  \begin{tabular}{@{}lccccccc@{}}
\toprule
Tokens (B) & 1 & 2 & 5 & 10 & 20 & 30 & 40 \\
\midrule
OCR, linear & $+$0.5 & $+$1.0 & $+$0.1 & $+$1.9 & $+$1.1 & $+$2.5 & $+$3.4 \\
OCR, frozen & $+$2.3 & $+$11.7 & $+$20.2 & $+$22.7 & $+$26.3 & $+$27.0 & $+$27.5 \\
GUI, linear & $+$0.1 & $+$0.3 & $+$1.9 & $+$3.8 & $+$8.3 & $+$9.4 & $+$10.8 \\
GUI, frozen & $+$1.4 & $+$1.7 & $+$14.4 & $+$32.2 & $+$41.9 & $+$45.6 & $+$46.7 \\
\bottomrule
\end{tabular}

\end{table}

\paragraph{Questions and answers.}
The multimodal loss scores every text token, questions included. \Cref{tab:supp_split} splits $\vigtot$ of the two 323M models between the tokens of the questions and those of the answers. On OCR images two thirds of the gain of the frozen-encoder condition falls on answer tokens. On general questions and STEM figures 83\% to 88\% of the gain falls on the question tokens, which make up 66\% to 70\% of the scored tokens. On GUI screenshots the frozen-encoder condition has 85\% of its gain on the questions, which quote the text on the screen that the action refers to. The image lowers the loss in large part by making the question predictable, and a model that reads the image predicts the quoted text.

\begin{table}[h]
  \caption{\textbf{Where the image gain falls.} Share of $\vigtot$ that falls on question tokens (gain) for the 323M decoder with the linear embedding and with the frozen encoder at 40B tokens, and share of question tokens among the scored tokens. Captions have no questions.}
  \label{tab:supp_split}
  \centering
  \footnotesize
  \setlength{\tabcolsep}{3pt}
  \begin{tabular}{@{}lcccccc@{}}
\toprule
 & Caption & Chart & General & OCR & STEM & GUI \\
\midrule
Gain, linear & 0\% & 75\% & 88\% & 59\% & 88\% & 41\% \\
Gain, frozen & 0\% & 68\% & 86\% & 33\% & 83\% & 85\% \\
Scored tokens & 0\% & 49\% & 70\% & 38\% & 66\% & 43\% \\
\bottomrule
\end{tabular}

\end{table}

\section{Growth of layout information}
\label{supp:growth}

\Cref{tab:supp_growth} fits $\viglay\propto D^{\beta}$ to each encoder-free run, with $D$ the number of training tokens. The exponent lies between 0.45 and 0.64 on captions and charts and between 0.69 and 0.76 on OCR images, and it does not order with model size. The two baseline seeds at 158M agree to 0.04 (0.58 and 0.55 on captions, 0.44 and 0.40 on charts, 0.74 and 0.73 on OCR images, fitted over 5B to 20B tokens). On general questions, STEM figures and GUI screenshots $\viglay$ stays close to zero over the first half of training and a power law does not describe it. The exponent is a description of the bootstrap phase. If the growth of the 323M decoder continued at this rate, it would reach the $\viglay$ that the frozen-encoder condition has at 40B tokens after about 130B tokens on charts, 450B on captions and more than 900B on OCR images. At our mixture 130B to 900B tokens are 0.2 to 1.3 billion image presentations, or 10 to 65 passes over our 19.8M samples, and NEO and Mono-InternVL train on several hundred million distinct image-text examples. The gap that remains at 40B tokens is therefore large on the scale of smooth growth. Whether it closes by such growth or in a faster regime, as the loss drop reported in Chen \etal~\cite{chen2026howfar} suggests, is open.

\begin{table*}[t]
  \caption{\textbf{Growth exponent of layout information.} Least-squares fit of $\log\viglay$ against $\log D$ over 5B to 40B tokens for the four encoder-free sizes. In parentheses the range over fits that start at 3B, 5B and 10B tokens.}
  \label{tab:supp_growth}
  \centering
  \small
  \begin{tabular}{@{}lcccc@{}}
\toprule
$\beta$ & 158M & 323M & 633M & 1.04B \\
\midrule
Caption & 0.57 (0.53 to 0.58) & 0.47 (0.47 to 0.48) & 0.53 (0.50 to 0.55) & 0.53 (0.47 to 0.57) \\
Chart & 0.49 (0.42 to 0.55) & 0.47 (0.47 to 0.54) & 0.45 (0.45 to 0.47) & 0.64 (0.64 to 0.67) \\
OCR & 0.76 (0.76 to 0.80) & 0.69 (0.69 to 0.76) & 0.69 (0.69 to 0.76) & 0.73 (0.63 to 0.90) \\
\bottomrule
\end{tabular}

\end{table*}

\section{Digitization and refits}
\label{supp:digitize}

\paragraph{What a comparison of decoder compute counts.}
Chen \etal~\cite{chen2026howfar} note that their compute-optimal models have passed the sharp loss drop that ends the slow early phase. In their loss curves, which we digitized, the drop lies at 28B to 32B total tokens for every model from 6.1B total parameters up (\cref{tab:supp_drop}), so the compute-optimal runs spend 22\% to 58\% of their multimodal compute before it. A compute law fitted to these runs includes this entry cost. In a refit the exponent advantage of encoder-free models moves from 0.078 to 0.055 when a third of it is counted as sunk and to 0.002 when all of it is (\cref{tab:supp_refit}). On the other side, the pretrained prior has its own cost, on the order of $3\times10^{22}$ FLOPs for SigLIP~2 so400m (Appendix~\ref{supp:siglipcost}), more than the reported crossover. It is paid once and reused by every model built on the encoder, and a comparison of decoder compute leaves it out. The rest of this section documents the digitization, the location of the drop and the refit.

\paragraph{Digitization.}
The four figures of Chen \etal~\cite{chen2026howfar} that carry the scaling data (IsoFLOP profiles and allocation, equal-loss compute laws, the multimodal loss transition and the envelope estimator) are matplotlib vector PDFs in the arXiv source. Axes were calibrated from the embedded tick labels and tick marks with zero residual, and every marker is a vector glyph whose center is the data point. Checks against numbers printed in Chen \etal~\cite{chen2026howfar}: the layer-12 attention masses 0.2171 and 0.6450 (reported 0.217 and 0.645), the observed efficiency gains 0.9853 / 0.4703 and 0.9974 / 0.7891 (reported 0.985 / 0.470 and 0.997 / 0.789), and the crossover marker at $6.14\times10^{21}$ FLOPs (reported $6.1\times10^{21}$). The IsoFLOP points of the two objectives were read off at budgets $2\times10^{19}$ to $2\times10^{20}$ FLOPs (text) and $10^{20}$ to $10^{21}$ FLOPs (multimodal). Refitting with their pipeline on the digitized vertices reproduces every reported exponent to within 0.2\%, the crossover, and the Appendix B.4 bootstrap intervals to within 0.002.

\paragraph{The loss drop.}
\Cref{tab:supp_drop} locates the multimodal loss drop of the encoder-free runs from the digitized loss-transition figure with a smoothed-slope detector. For every run from 6.1B to 44B total parameters the drop lies at 28B to 32B total tokens, a spread smaller than the checkpoint spacing. Below 6.1B it moves later as the model shrinks, and the 1.1B run shows no completed drop within its 95B tokens. No encoder-based curve has a drop. Every multimodal IsoFLOP point lies after the drop of its run. The pre-drop slope is $-0.004$ to $-0.016$ nats per billion tokens, so the pre-drop phase is slow but not flat.

\begin{table}[h]
  \caption{\textbf{Loss drop of the encoder-free runs of Chen \etal~\cite{chen2026howfar}}, located on the digitized loss-transition figure. Active parameters for 2.3B, 3.4B and 4.5B are interpolated on a log-log scale between the sizes stated in the paper.}
  \label{tab:supp_drop}
  \centering
  \footnotesize
  \setlength{\tabcolsep}{3.5pt}
  \begin{tabular}{@{}rrrr@{}}
    \toprule
    Total params & Active & Drop (B tokens) & Size (nats) \\
    \midrule
    1.1B & 71M & none within 95B & \\
    2.3B & 158M & 45.6 & 0.18 \\
    3.4B & 232M & 35.4 & 0.18 \\
    4.5B & 305M & 35.8 & 0.20 \\
    6.1B & 425M & 31.1 & 0.26 \\
    8B & 577M & 29.5 & 0.29 \\
    13B & 922M & 28.3 & 0.28 \\
    16B & 1.0B & 31.5 & 0.29 \\
    22B & 1.6B & 28.3 & 0.30 \\
    33B & 2.2B & 29.4 & 0.39 \\
    44B & 2.4B & 31.5 & 0.40 \\
    \bottomrule
  \end{tabular}
\end{table}

\paragraph{Sunk-cost refit.}
The refit treats a fixed number of training tokens before the loss drop as sunk compute for every encoder-free run and refits their compute law, $\mathcal{L}^{*}(C)=E+KC^{-\gamma}$, to the remaining compute. The encoder-based runs enter with their full compute. As in the original fit, the irreducible loss $E$ is shared between the two systems and fitted jointly. This lowers $E$ as more compute is sunk, and $\gamma_{\text{based}}$ follows $E$ although the encoder-based data do not change. \Cref{tab:supp_refit} therefore repeats the refit with $E$ held at the published value of 0.403, which our refit of the original data reproduces as 0.404. $\gamma_{\text{based}}$ then stays at 0.299, and $\Delta\gamma$ and the crossover behave as in the joint refit: the advantage survives with 10B tokens sunk and is gone with 29.5B. A third variant with a separate $E$ for each system is not identified by six frontier points per system. It returns $E=0.196$ for the encoder-based law and 0.37 to 0.45 for the encoder-free law with up to 29.5B tokens sunk, floors under which the two frontiers do not cross, so we keep the shared floor of the original analysis.

\begin{table}[h]
  \caption{\textbf{Sunk-cost refit under two treatments of the irreducible loss.} Sunk compute is removed from the encoder-free runs only. Crossover in FLOPs.}
  \label{tab:supp_refit}
  \centering
  \footnotesize
  \setlength{\tabcolsep}{3.5pt}
  \begin{tabular}{@{}lccccc@{}}
    \toprule
    Sunk tokens & $E$ & $\gamma_{\text{free}}$ & $\gamma_{\text{based}}$ & $\Delta\gamma$ & Crossover \\
    \midrule
    \multicolumn{6}{@{}l}{\textit{$E$ shared and refitted (values quoted above)}} \\
    0 (original fit) & 0.404 & 0.378 & 0.300 & $+0.078$ & $6.1\times10^{21}$ \\
    10B   & 0.396 & 0.348 & 0.293 & $+0.055$ & $9.5\times10^{21}$ \\
    20B   & 0.377 & 0.307 & 0.278 & $+0.029$ & $3.0\times10^{22}$ \\
    29.5B & 0.331 & 0.250 & 0.248 & $+0.002$ & none \\
    40B   & 0.134 & 0.150 & 0.170 & $-0.020$ & none \\
    \addlinespace[2pt]
    \multicolumn{6}{@{}l}{\textit{$E$ held at the published value}} \\
    0 & 0.403 & 0.377 & 0.299 & $+0.078$ & $6.1\times10^{21}$ \\
    10B   & 0.403 & 0.354 & 0.299 & $+0.055$ & $1.0\times10^{22}$ \\
    20B   & 0.403 & 0.325 & 0.299 & $+0.026$ & $6.2\times10^{22}$ \\
    29.5B & 0.403 & 0.290 & 0.299 & $-0.009$ & none \\
    40B   & 0.403 & 0.237 & 0.299 & $-0.062$ & none \\
    \bottomrule
  \end{tabular}
\end{table}

\noindent The compute-optimal runs spend 58\% of their multimodal compute before the drop at $10^{20}$ FLOPs and 22\% at $10^{21}$. The bootstrap for the fully sunk row follows their paired residual bootstrap and gives $\gamma_{\text{free}}$ in $[0.209, 0.288]$, $\gamma_{\text{based}}$ in $[0.200, 0.295]$ and $\Delta\gamma$ in $[-0.010, 0.014]$ as central 80\% intervals, with a finite crossover in 30\% of the replicates. A two-phase parametric fit on the IsoFLOP points reproduces the reported allocation shift when nothing is sunk and becomes degenerate once 29.5B tokens are sunk, because the points right after the drop have as few as 2B post-drop tokens. A re-implementation of their envelope estimator, matching its digitized points to $10^{-5}$ in log model size, changes the exponents by less than 0.01 when the pre-drop segments are removed, so the frontier estimates themselves are not contaminated by pre-drop points. The sunk cost enters only through the compute already spent inside the post-drop runs.

\section{Training details}
\label{supp:training}

\paragraph{Decoder.}
\Cref{tab:ladder} lists the four sizes. All decoders use a head dimension of 64, SwiGLU feed-forward layers of width 2{,}816, 3{,}456, 4{,}864 and 5{,}632 for the four sizes, RMSNorm, rotary position embeddings, tied input and output embeddings over the Qwen2.5 vocabulary (an embedding table of 151{,}936 entries), and document masking with bidirectional attention among the tokens of one image. Sequences hold 4{,}096 tokens (8{,}192 for the 16-pixel condition, which keeps 0.52M tokens per step by halving the number of sequences). Warmup is 2{,}000 steps for every run of the ladder, which is 1.05B tokens at 158M. The 16-pixel condition and the runs with batch 512 and batch 32 keep these 1.05B warmup tokens and the peak learning rate of the 158M baseline, $1.5\times10^{-3}$, so the batch-512 run warms up over 500 steps and the batch-32 run over 8{,}000. Checkpoints are saved every billion tokens and evaluated every 500M tokens on text and multimodal validation sets.

\begin{table}[h]
  \caption{\textbf{Encoder-free ladder.} Non-embedding parameters, width, depth, batch in 4{,}096-token sequences, peak learning rate and training tokens.}
  \label{tab:ladder}
  \centering
  \small
  \begin{tabular}{@{}lccccc@{}}
    \toprule
    Params & $d$ & $L$ & Batch (seq.) & Peak lr & Tokens \\
    \midrule
    158M  & 1024 & 12 & 128 & $1.5\times10^{-3}$ & 40B \\
    323M  & 1280 & 16 & 256 & $1.0\times10^{-3}$ & 40B \\
    633M  & 1792 & 16 & 256 & $7\times10^{-4}$   & 40B \\
    1.04B & 2048 & 20 & 512 & $5\times10^{-4}$   & 40B \\
    \bottomrule
  \end{tabular}
\end{table}

\paragraph{Mixtures.}
The image-text half draws one sample at a time from a source chosen with probability proportional to its number of shards times a per-source weight, which is 1 for every source in the default mixture. Shards hold about 2{,}000 samples except for Docmatix, whose shards hold 863 after multi-image samples are removed, so Docmatix is drawn about 2.3 times as often as its size implies. By tokens, the default mixture is 42\% captions, 21\% Docmatix, 16\% LLaVA-OneVision, 13\% the Cauldron, 6\% Aguvis and 2\% chart and STEM sources, and by image type 44\% captions, 28\% OCR, 11\% charts, 7\% STEM, 7\% GUI and 4\% general questions. Over the 20B image-text tokens of a 40B run every source is seen 2.8 to 3.2 times, except Docmatix, which is seen 6.5 times. The document-heavy mixture raises the weights of the non-caption sources by factors of 4.4 to 8.6. Its token shares are 43\% Docmatix, 19\% LLaVA-OneVision, 15\% the Cauldron, 8\% Aguvis, 4\% chart and STEM sources and 11\% captions, and 52\% of its tokens are OCR images. Over its 10B image-text tokens Docmatix is seen 6.8 times, against 3.3 times for the default mixture on the same schedule, and no other source more than 3.3 times. An image carries 176 text tokens on average, against 86 in the default mixture and 21 in the caption-only mixture. With 16-pixel patches an image takes 1{,}016 image tokens instead of 254, so the run sees 3.2 times fewer images per token. Docmatix carries about 361 text tokens per image, Aguvis about 320, the conversation sources 138 to 241 and captions 11 to 24. The caption-only mixture draws the image-text half from CC12M and CC3M alone.

\paragraph{Front ends.}
The linear embedding maps each flattened $32\times32$ pixel patch by LayerNorm, a linear layer and LayerNorm, adds factorized two-dimensional position embeddings, and normalizes and projects the result to the model width. The frozen-encoder condition uses the so400m NaFlex encoder (412.6M frozen parameters) with a $2\times2$ convolutional pooling and an MLP projector (6.8M trainable parameters at 158M and 9.2M at 323M). The random-init ViT condition instantiates the same architecture (27 layers, width 1{,}152, 16 heads, feed-forward width 4{,}304) with random weights, trains it jointly with gradient checkpointing, and triples the compute of the run. It was trained on a 20B schedule to 3.8B tokens, continued on a 10B schedule, and stopped after 8.1B tokens under a decision rule declared in advance: the continuation would be kept only if its OCR or GUI layout information at 6B tokens reached 0.10 nats, and both values stayed below 0.01.

\paragraph{Hardware.}
Runs were trained with PyTorch, FlexAttention and bf16 on nodes with eight H200 GPUs, nodes with four L40S GPUs and a standalone four-GPU server, with gradient accumulation set so that the batch per optimizer step is independent of the node type. Evaluation of one checkpoint under the four conditions takes under a minute for a 158M decoder on one L40S.

\section{Compute of the SigLIP 2 encoder}
\label{supp:siglipcost}

SigLIP~2~\cite{tschannen2025siglip2} trains the so400m image tower (about 400M parameters) together with a text tower of the same size for 40B image-text examples at batch size 32k. At 256 pixels with 16-pixel patches the image tower processes 256 tokens per example and the text tower 64, so one forward pass costs about $2\times4\times10^{8}\times(256+64)\approx2.6\times10^{11}$ FLOPs, and forward plus backward about $7.7\times10^{11}$. Over 40B examples this is $3.1\times10^{22}$ FLOPs before the decoder-based losses and the later multi-resolution stages, which only add to it. The estimate quoted in Appendix~\ref{supp:digitize}, on the order of $3\times10^{22}$ FLOPs, is therefore a lower bound on the compute behind the frozen prior.

\section{Visual information gain of public models}
\label{supp:public}

\paragraph{Protocol.}
The estimator needs only the loss of a model on a text under three versions of the image, so it applies to released checkpoints. We score the validation samples of \cref{tab:supp_val}, the same images and the same text as for our own runs, with 19 public checkpoints. Seven are scored through the processor and model classes of their release and are described first. The other twelve are released training stages of two encoder-free models with their own modeling code (last paragraph of this section). Gemma~4 12B~\cite{gemma2026gemma4} feeds pixel patches to the decoder without a visual encoder and is scored in its pretrained and its instruction-tuned release. Gemma~4 26B-A4B, Qwen3-VL-8B~\cite{bai2025qwen3vl}, Qwen2.5-VL-7B~\cite{bai2025qwen25vl}, InternVL3-8B~\cite{zhu2025internvl3} and LLaVA-1.5-7B~\cite{liu2024llava15} are built on visual encoders. Each image passes through the processor of the model at its default settings, and for Gemma~4 12B also at a budget of 1{,}120 image tokens. The shuffled image permutes the $32\times32$ pixel blocks of the image with a fixed seed per sample, identical for every model, and the uniform image sets every pixel to 128. In the raw format the text follows the image tokens directly, as in our runs. In the chat format the image is the user turn of the chat template of the model and the text is the reply. Only the tokens of the text are scored. Tokenizers differ between models, so the summed loss of each model is divided by the number of scored tokens of our own evaluation of the same samples (\cref{tab:supp_val}), which puts every model in the unit of the main text.

\paragraph{Image types that can be compared.}
The comparison presupposes that the scored text lies in the training distribution of the model, so that the loss with the uniform image measures language modeling. On OCR images this holds. The seven checkpoints reach 1.48 to 2.01 nats with the uniform image, against 1.57 to 1.75 for our decoders. Docmatix is public and may be among their training data. A loss with the uniform image in the range of our decoders, which have not seen these documents, shows that the text is not memorized. On GUI screenshots they reach 1.70 to 2.26 against 1.38 to 1.46, so the action format of our GUI data is less familiar to them and the comparison is weaker. On charts, captions, STEM figures and general questions the loss of the seven checkpoints with the real image is 1.2 to 2.8 times that of our 323M decoder per character, and for Qwen2.5-VL, Qwen3-VL and InternVL3 the real image raises the loss on general questions. \Cref{tab:supp_public} therefore reports OCR images and GUI screenshots.

\begin{table*}[t]
  \caption{\textbf{Public models on the shared validation samples.} Loss with the real image $\Lreal$, image-blind loss $\Lblank$, layout information $\viglay$ and global information $\vigglob$ on OCR images and GUI screenshots, in nats per scored token of our evaluation. Image tokens are the mean number per OCR image. Our decoders are means over the last three checkpoints. The last block lists the released training stages of two encoder-free models.}
  \label{tab:supp_public}
  \centering
  \small
  \setlength{\tabcolsep}{2.2pt}
  \begin{tabular}{@{}lllrcccccccc@{}}
\toprule
 & & & & \multicolumn{4}{c}{OCR} & \multicolumn{4}{c}{GUI} \\
\cmidrule(lr){5-8}\cmidrule(l){9-12}
Model & Encoder & Format & Image tokens & $\Lreal$ & $\Lblank$ & $\viglay$ & $\vigglob$ & $\Lreal$ & $\Lblank$ & $\viglay$ & $\vigglob$ \\
\midrule
\multicolumn{12}{@{}l}{\textit{Our decoders, 40B tokens}} \\
158M, linear embedding & none & raw & 768 & 1.611 & 1.717 & 0.068 & 0.038 & 1.410 & 1.454 & 0.054 & $-$0.010 \\
323M, linear embedding & none & raw & 768 & 1.518 & 1.622 & 0.064 & 0.040 & 1.377 & 1.416 & 0.030 & 0.008 \\
633M, linear embedding & none & raw & 768 & 1.463 & 1.573 & 0.063 & 0.047 & 1.350 & 1.395 & 0.036 & 0.009 \\
1.04B, linear embedding & none & raw & 768 & 1.447 & 1.566 & 0.042 & 0.077 & 1.340 & 1.380 & 0.020 & 0.021 \\
323M, frozen SigLIP~2 & frozen & raw & 768 & 0.921 & 1.748 & 0.698 & 0.130 & 0.701 & 1.457 & 0.702 & 0.054 \\
\addlinespace[2pt]
\multicolumn{12}{@{}l}{\textit{Public models}} \\
Gemma~4 12B~\cite{gemma2026gemma4} & none & raw & 259 & 0.931 & 1.494 & 0.525 & 0.038 & 1.166 & 1.739 & 0.524 & 0.049 \\
Gemma~4 12B & none & raw & 1{,}082 & 0.787 & 1.478 & 0.622 & 0.068 & 1.125 & 1.740 & 0.483 & 0.132 \\
Gemma~4 12B-it & none & chat & 259 & 1.171 & 1.785 & 0.616 & $-$0.002 & 1.304 & 2.038 & 0.729 & 0.005 \\
Gemma~4 12B-it & none & chat & 1{,}082 & 0.885 & 1.744 & 0.758 & 0.100 & 1.072 & 1.977 & 0.778 & 0.127 \\
Gemma~4 26B-A4B-it & yes & chat & 259 & 1.184 & 2.011 & 0.820 & 0.007 & 1.397 & 2.227 & 0.843 & $-$0.013 \\
Qwen3-VL-8B~\cite{bai2025qwen3vl} & yes & chat & 768 & 0.910 & 1.701 & 0.702 & 0.090 & 1.272 & 2.010 & 0.660 & 0.078 \\
Qwen2.5-VL-7B~\cite{bai2025qwen25vl} & yes & chat & 1{,}018 & 0.804 & 1.608 & 0.681 & 0.124 & 1.243 & 2.049 & 0.542 & 0.264 \\
Qwen2.5-VL-7B & yes & raw & 1{,}018 & 0.915 & 1.621 & 0.645 & 0.060 & 1.505 & 2.257 & 0.572 & 0.180 \\
InternVL3-8B~\cite{zhu2025internvl3} & yes & chat & 3{,}288 & 0.842 & 1.557 & 0.675 & 0.039 & 1.038 & 1.700 & 0.684 & $-$0.022 \\
LLaVA-1.5-7B~\cite{liu2024llava15} & yes & chat & 576 & 1.417 & 1.629 & 0.194 & 0.018 & 1.771 & 2.014 & 0.206 & 0.037 \\
\addlinespace[2pt]
\multicolumn{12}{@{}l}{\textit{Public encoder-free models with released training stages}} \\
NEO-2B, pre-training~\cite{diao2025neo} & none & chat & 768 & 1.883 & 2.432 & 0.506 & 0.043 & 2.097 & 2.713 & 0.487 & 0.129 \\
NEO-2B, mid-training & none & chat & 768 & 1.002 & 1.668 & 0.648 & 0.019 & 1.315 & 1.972 & 0.627 & 0.030 \\
NEO-2B, fine-tuning & none & chat & 768 & 0.993 & 1.676 & 0.665 & 0.018 & 1.392 & 2.027 & 0.612 & 0.023 \\
NEO1.5-2B, fine-tuning & none & chat & 768 & 0.978 & 1.752 & 0.724 & 0.051 & 1.177 & 1.959 & 0.693 & 0.088 \\
NEO-9B, pre-training & none & chat & 768 & 1.299 & 1.653 & 0.353 & 0.001 & 1.549 & 1.969 & 0.410 & 0.009 \\
NEO-9B, mid-training & none & chat & 768 & 0.907 & 1.512 & 0.569 & 0.035 & 1.235 & 1.855 & 0.601 & 0.019 \\
NEO-9B, fine-tuning & none & chat & 768 & 0.912 & 1.535 & 0.595 & 0.029 & 1.253 & 1.901 & 0.618 & 0.030 \\
NEO1.5-9B, fine-tuning & none & chat & 768 & 0.891 & 1.588 & 0.673 & 0.025 & 1.074 & 1.786 & 0.734 & $-$0.022 \\
Mono-InternVL-2B, S1.1~\cite{luo2025monointernvl} & none & chat & 3{,}288 & 1.652 & 2.050 & 0.268 & 0.130 & 1.931 & 2.430 & 0.301 & 0.198 \\
Mono-InternVL-2B, S1.2 & none & chat & 3{,}288 & 1.457 & 1.891 & 0.410 & 0.025 & 1.942 & 2.252 & 0.295 & 0.015 \\
Mono-InternVL-2B, S1.3 & none & chat & 3{,}288 & 1.215 & 1.944 & 0.674 & 0.055 & 1.728 & 2.398 & 0.580 & 0.090 \\
Mono-InternVL-2B, final & none & chat & 3{,}288 & 1.160 & 1.916 & 0.715 & 0.041 & 1.721 & 2.386 & 0.585 & 0.080 \\
\bottomrule
\end{tabular}

\end{table*}

\paragraph{Results.}
On OCR images each of the seven checkpoints except LLaVA-1.5 has between 0.53 and 0.82 nats of layout information, and layout information makes up 85\% to 100\% of the image gain of all seven. The frozen-encoder condition of our 323M decoder lies inside this range, at 0.70 after 40B tokens, and our encoder-free decoders lie an order of magnitude below it, at 0.04 to 0.07. LLaVA-1.5, whose CLIP encoder sees the image at 336 pixels, reaches 0.19. The encoder-free Gemma~4 12B reaches 0.53 to 0.76 depending on release, format and image-token budget, which overlaps the range of the encoder-based models (0.65 to 0.82). At the same budget of about 260 image tokens and in the same format, the encoder-based Gemma~4 26B-A4B is 0.20 nats above the encoder-free 12B (paired bootstrap 95\% interval 0.16 to 0.25), in line with the document-parsing results reported for the encoder-free and the encoder-based Gemma~4 models~\cite{gemma2026gemma4}. The two models also differ in size. GUI screenshots show the same picture, with 0.48 to 0.84 nats for the checkpoints other than LLaVA-1.5, 0.70 for our frozen-encoder condition and 0.02 to 0.05 for our encoder-free decoders. Bootstrap 95\% intervals over samples are about $\pm0.08$ nats on OCR images and $\pm0.10$ to $\pm0.15$ on GUI screenshots.

\paragraph{Training stages of encoder-free models.}
Two further encoder-free models are released at several stages of their training, which shows the same quantity over training in models at production scale. NEO~\cite{diao2025neo} is released after pre-training, mid-training and supervised fine-tuning, at 2B and 9B parameters and in a later version, NEO1.5. Mono-InternVL-2B~\cite{luo2025monointernvl} is released after the three stages S1.1, S1.2 and S1.3 of its visual pre-training and as the final model. Both use their own modeling code. We follow the image preprocessing of their official evaluation code, one image token per $32\times32$ pixels for NEO and at most 12 tiles of 448 pixels for Mono-InternVL, and score OCR images and GUI screenshots in the chat format (last block of \cref{tab:supp_public}). On OCR images layout information rises from 0.27 nats after the first stage of Mono-InternVL-2B to 0.41, 0.67 and 0.72 after the later ones, from 0.51 to 0.65 and 0.67 over the three stages of NEO-2B, and from 0.35 to 0.57 and 0.60 for NEO-9B. The NEO1.5 releases reach 0.72 and 0.67. Every final model lies between 0.53 and 0.82 nats, the range of the seven checkpoints above. The loss with the uniform image changes little over these stages, from 2.05 to 1.92 nats for Mono-InternVL-2B and from 1.65 to 1.54 for NEO-9B, so the rise does not come from growing familiarity with the text. For NEO-2B it falls from 2.43 to 1.67 between the first two stages, and we do not use this pair in the main text. The stages of both models differ in their data and in which weights are trained, so the rise shows that layout information is acquired during training and does not isolate what brings it. The pre-training release of NEO-2B carries a configuration entry for its attention pattern that the released code does not read, so its row may differ from the model as trained.

\section{Full decomposition tables}
\label{supp:tables}

The following tables list $\Lreal$, $\Lblank$, $\vigglob$ and $\viglay$ for every run and image type at single checkpoints of 1, 2, 5, 10, 20 and 40B tokens, where the run reaches them. The loss with the shuffled image follows as $\Lperm=\Lreal+\viglay$.

\begin{table*}[p]\centering\small
\caption*{\textbf{158M, linear embedding, 40B schedule (ladder baseline).} Nats per text token, single checkpoints.}
\begin{tabular}{@{}llcccccc@{}}
\toprule
Image type & quantity & 1B & 2B & 5B & 10B & 20B & 40B \\
\midrule
Caption & $\Lreal$ & 3.776 & 3.454 & 3.223 & 3.115 & 2.963 & 2.749 \\
 & $\Lblank$ & 4.200 & 3.813 & 3.646 & 3.587 & 3.493 & 3.344 \\
 & $\vigglob$ & 0.399 & 0.296 & 0.338 & 0.350 & 0.346 & 0.335 \\
 & $\viglay$ & 0.025 & 0.063 & 0.085 & 0.122 & 0.184 & 0.261 \\
\addlinespace[1pt]
Chart & $\Lreal$ & 2.018 & 1.778 & 1.759 & 1.619 & 1.432 & 1.290 \\
 & $\Lblank$ & 2.371 & 2.161 & 2.256 & 2.051 & 1.939 & 1.849 \\
 & $\vigglob$ & 0.271 & 0.270 & 0.326 & 0.249 & 0.201 & 0.152 \\
 & $\viglay$ & 0.082 & 0.113 & 0.170 & 0.183 & 0.306 & 0.406 \\
\addlinespace[1pt]
General & $\Lreal$ & 1.593 & 1.433 & 1.343 & 1.336 & 1.231 & 1.146 \\
 & $\Lblank$ & 1.693 & 1.502 & 1.448 & 1.441 & 1.420 & 1.344 \\
 & $\vigglob$ & 0.100 & 0.067 & 0.098 & 0.096 & 0.159 & 0.130 \\
 & $\viglay$ & 0.000 & 0.003 & 0.007 & 0.009 & 0.031 & 0.068 \\
\addlinespace[1pt]
OCR & $\Lreal$ & 2.285 & 2.034 & 1.877 & 1.804 & 1.722 & 1.609 \\
 & $\Lblank$ & 2.327 & 2.059 & 1.914 & 1.856 & 1.792 & 1.716 \\
 & $\vigglob$ & 0.039 & 0.020 & 0.024 & 0.028 & 0.032 & 0.040 \\
 & $\viglay$ & 0.003 & 0.005 & 0.013 & 0.024 & 0.038 & 0.068 \\
\addlinespace[1pt]
STEM & $\Lreal$ & 1.686 & 1.405 & 1.266 & 1.189 & 1.101 & 0.991 \\
 & $\Lblank$ & 1.756 & 1.457 & 1.332 & 1.258 & 1.176 & 1.074 \\
 & $\vigglob$ & 0.071 & 0.049 & 0.058 & 0.061 & 0.062 & 0.061 \\
 & $\viglay$ & 0.000 & 0.003 & 0.007 & 0.007 & 0.013 & 0.022 \\
\addlinespace[1pt]
GUI & $\Lreal$ & 1.721 & 1.609 & 1.535 & 1.511 & 1.477 & 1.412 \\
 & $\Lblank$ & 1.750 & 1.630 & 1.556 & 1.554 & 1.516 & 1.459 \\
 & $\vigglob$ & 0.029 & 0.022 & 0.017 & 0.033 & 0.021 & $-$0.008 \\
 & $\viglay$ & 0.000 & $-$0.001 & 0.004 & 0.010 & 0.018 & 0.055 \\
\addlinespace[1pt]
\bottomrule
\end{tabular}
\end{table*}

\begin{table*}[p]\centering\small
\caption*{\textbf{323M, linear embedding, 40B schedule.} Nats per text token, single checkpoints.}
\begin{tabular}{@{}llcccccc@{}}
\toprule
Image type & quantity & 1B & 2B & 5B & 10B & 20B & 40B \\
\midrule
Caption & $\Lreal$ & 3.867 & 3.544 & 3.154 & 2.987 & 2.801 & 2.605 \\
 & $\Lblank$ & 4.186 & 3.945 & 3.609 & 3.475 & 3.331 & 3.322 \\
 & $\vigglob$ & 0.299 & 0.350 & 0.369 & 0.365 & 0.359 & 0.485 \\
 & $\viglay$ & 0.020 & 0.051 & 0.086 & 0.122 & 0.171 & 0.232 \\
\addlinespace[1pt]
Chart & $\Lreal$ & 2.087 & 1.809 & 1.678 & 1.510 & 1.291 & 1.153 \\
 & $\Lblank$ & 2.562 & 2.215 & 2.109 & 1.989 & 1.912 & 1.848 \\
 & $\vigglob$ & 0.396 & 0.277 & 0.233 & 0.236 & 0.281 & 0.211 \\
 & $\viglay$ & 0.079 & 0.129 & 0.198 & 0.242 & 0.340 & 0.484 \\
\addlinespace[1pt]
General & $\Lreal$ & 1.638 & 1.460 & 1.312 & 1.271 & 1.168 & 1.080 \\
 & $\Lblank$ & 1.743 & 1.569 & 1.503 & 1.448 & 1.381 & 1.369 \\
 & $\vigglob$ & 0.104 & 0.104 & 0.169 & 0.150 & 0.186 & 0.217 \\
 & $\viglay$ & 0.001 & 0.006 & 0.022 & 0.027 & 0.028 & 0.072 \\
\addlinespace[1pt]
OCR & $\Lreal$ & 2.352 & 2.089 & 1.819 & 1.705 & 1.618 & 1.517 \\
 & $\Lblank$ & 2.392 & 2.120 & 1.860 & 1.769 & 1.692 & 1.619 \\
 & $\vigglob$ & 0.038 & 0.027 & 0.026 & 0.041 & 0.039 & 0.037 \\
 & $\viglay$ & 0.001 & 0.004 & 0.015 & 0.023 & 0.035 & 0.065 \\
\addlinespace[1pt]
STEM & $\Lreal$ & 1.753 & 1.456 & 1.198 & 1.098 & 0.996 & 0.897 \\
 & $\Lblank$ & 1.824 & 1.524 & 1.259 & 1.176 & 1.079 & 0.998 \\
 & $\vigglob$ & 0.072 & 0.068 & 0.055 & 0.071 & 0.066 & 0.077 \\
 & $\viglay$ & $-$0.001 & 0.001 & 0.005 & 0.007 & 0.017 & 0.023 \\
\addlinespace[1pt]
GUI & $\Lreal$ & 1.769 & 1.644 & 1.511 & 1.471 & 1.434 & 1.378 \\
 & $\Lblank$ & 1.804 & 1.663 & 1.537 & 1.501 & 1.466 & 1.419 \\
 & $\vigglob$ & 0.035 & 0.019 & 0.025 & 0.023 & 0.017 & 0.010 \\
 & $\viglay$ & 0.001 & 0.000 & 0.000 & 0.007 & 0.014 & 0.031 \\
\addlinespace[1pt]
\bottomrule
\end{tabular}
\end{table*}

\begin{table*}[p]\centering\small
\caption*{\textbf{633M, linear embedding, 40B schedule.} Nats per text token, single checkpoints.}
\begin{tabular}{@{}llcccccc@{}}
\toprule
Image type & quantity & 1B & 2B & 5B & 10B & 20B & 40B \\
\midrule
Caption & $\Lreal$ & 3.844 & 3.507 & 3.090 & 2.888 & 2.684 & 2.474 \\
 & $\Lblank$ & 4.222 & 3.950 & 3.567 & 3.403 & 3.310 & 3.273 \\
 & $\vigglob$ & 0.357 & 0.404 & 0.397 & 0.393 & 0.438 & 0.539 \\
 & $\viglay$ & 0.022 & 0.040 & 0.080 & 0.123 & 0.188 & 0.261 \\
\addlinespace[1pt]
Chart & $\Lreal$ & 2.039 & 1.783 & 1.648 & 1.450 & 1.219 & 1.050 \\
 & $\Lblank$ & 2.539 & 2.293 & 2.114 & 1.950 & 1.894 & 1.817 \\
 & $\vigglob$ & 0.420 & 0.388 & 0.272 & 0.242 & 0.310 & 0.264 \\
 & $\viglay$ & 0.080 & 0.122 & 0.193 & 0.258 & 0.366 & 0.503 \\
\addlinespace[1pt]
General & $\Lreal$ & 1.621 & 1.447 & 1.292 & 1.236 & 1.135 & 1.051 \\
 & $\Lblank$ & 1.768 & 1.559 & 1.481 & 1.426 & 1.357 & 1.351 \\
 & $\vigglob$ & 0.145 & 0.111 & 0.184 & 0.180 & 0.203 & 0.265 \\
 & $\viglay$ & 0.001 & 0.001 & 0.005 & 0.009 & 0.020 & 0.035 \\
\addlinespace[1pt]
OCR & $\Lreal$ & 2.340 & 2.055 & 1.781 & 1.651 & 1.554 & 1.462 \\
 & $\Lblank$ & 2.387 & 2.097 & 1.820 & 1.724 & 1.643 & 1.571 \\
 & $\vigglob$ & 0.044 & 0.037 & 0.024 & 0.049 & 0.051 & 0.047 \\
 & $\viglay$ & 0.002 & 0.005 & 0.015 & 0.024 & 0.038 & 0.061 \\
\addlinespace[1pt]
STEM & $\Lreal$ & 1.739 & 1.425 & 1.168 & 1.048 & 0.941 & 0.842 \\
 & $\Lblank$ & 1.822 & 1.493 & 1.242 & 1.126 & 1.028 & 0.955 \\
 & $\vigglob$ & 0.084 & 0.066 & 0.066 & 0.070 & 0.073 & 0.093 \\
 & $\viglay$ & $-$0.001 & 0.002 & 0.008 & 0.008 & 0.014 & 0.020 \\
\addlinespace[1pt]
GUI & $\Lreal$ & 1.766 & 1.625 & 1.492 & 1.449 & 1.404 & 1.351 \\
 & $\Lblank$ & 1.791 & 1.647 & 1.519 & 1.483 & 1.441 & 1.396 \\
 & $\vigglob$ & 0.026 & 0.022 & 0.027 & 0.031 & 0.021 & 0.009 \\
 & $\viglay$ & $-$0.001 & 0.000 & 0.000 & 0.004 & 0.016 & 0.036 \\
\addlinespace[1pt]
\bottomrule
\end{tabular}
\end{table*}

\begin{table*}[p]\centering\small
\caption*{\textbf{1.04B, linear embedding, 40B schedule.} Nats per text token, single checkpoints.}
\begin{tabular}{@{}llcccccc@{}}
\toprule
Image type & quantity & 1B & 2B & 5B & 10B & 20B & 40B \\
\midrule
Caption & $\Lreal$ & 4.622 & 3.719 & 3.215 & 2.923 & 2.651 & 2.423 \\
 & $\Lblank$ & 5.130 & 4.291 & 3.723 & 3.481 & 3.295 & 3.280 \\
 & $\vigglob$ & 0.503 & 0.549 & 0.448 & 0.448 & 0.488 & 0.650 \\
 & $\viglay$ & 0.005 & 0.022 & 0.060 & 0.111 & 0.157 & 0.207 \\
\addlinespace[1pt]
Chart & $\Lreal$ & 2.600 & 2.040 & 1.758 & 1.454 & 1.172 & 0.961 \\
 & $\Lblank$ & 3.350 & 2.483 & 2.201 & 2.032 & 1.861 & 1.825 \\
 & $\vigglob$ & 0.711 & 0.404 & 0.322 & 0.355 & 0.404 & 0.388 \\
 & $\viglay$ & 0.038 & 0.038 & 0.121 & 0.223 & 0.285 & 0.476 \\
\addlinespace[1pt]
General & $\Lreal$ & 1.946 & 1.583 & 1.338 & 1.237 & 1.123 & 1.044 \\
 & $\Lblank$ & 2.231 & 1.685 & 1.432 & 1.347 & 1.305 & 1.266 \\
 & $\vigglob$ & 0.284 & 0.101 & 0.092 & 0.103 & 0.170 & 0.199 \\
 & $\viglay$ & 0.001 & 0.000 & 0.002 & 0.006 & 0.012 & 0.023 \\
\addlinespace[1pt]
OCR & $\Lreal$ & 3.579 & 2.264 & 1.850 & 1.654 & 1.532 & 1.449 \\
 & $\Lblank$ & 3.739 & 2.343 & 1.901 & 1.717 & 1.639 & 1.563 \\
 & $\vigglob$ & 0.159 & 0.079 & 0.046 & 0.044 & 0.079 & 0.073 \\
 & $\viglay$ & 0.001 & 0.000 & 0.005 & 0.019 & 0.029 & 0.041 \\
\addlinespace[1pt]
STEM & $\Lreal$ & 2.947 & 1.700 & 1.244 & 1.049 & 0.918 & 0.818 \\
 & $\Lblank$ & 3.124 & 1.817 & 1.325 & 1.132 & 1.008 & 0.931 \\
 & $\vigglob$ & 0.177 & 0.117 & 0.079 & 0.081 & 0.083 & 0.097 \\
 & $\viglay$ & 0.000 & 0.000 & 0.002 & 0.002 & 0.008 & 0.015 \\
\addlinespace[1pt]
GUI & $\Lreal$ & 2.100 & 1.719 & 1.530 & 1.455 & 1.395 & 1.340 \\
 & $\Lblank$ & 2.065 & 1.787 & 1.554 & 1.496 & 1.430 & 1.382 \\
 & $\vigglob$ & $-$0.033 & 0.068 & 0.026 & 0.041 & 0.026 & 0.022 \\
 & $\viglay$ & $-$0.002 & 0.000 & $-$0.002 & 0.000 & 0.010 & 0.020 \\
\addlinespace[1pt]
\bottomrule
\end{tabular}
\end{table*}

\begin{table*}[p]\centering\small
\caption*{\textbf{323M, frozen SigLIP~2 encoder, 40B schedule.} Nats per text token, single checkpoints.}
\begin{tabular}{@{}llcccccc@{}}
\toprule
Image type & quantity & 1B & 2B & 5B & 10B & 20B & 40B \\
\midrule
Caption & $\Lreal$ & 3.436 & 2.974 & 2.582 & 2.425 & 2.288 & 2.117 \\
 & $\Lblank$ & 4.388 & 4.159 & 3.867 & 3.717 & 3.593 & 3.597 \\
 & $\vigglob$ & 0.531 & 0.585 & 0.573 & 0.587 & 0.586 & 0.750 \\
 & $\viglay$ & 0.421 & 0.601 & 0.712 & 0.705 & 0.719 & 0.731 \\
\addlinespace[1pt]
Chart & $\Lreal$ & 1.942 & 1.641 & 1.424 & 1.230 & 1.007 & 0.907 \\
 & $\Lblank$ & 2.637 & 2.341 & 2.242 & 2.181 & 2.153 & 2.186 \\
 & $\vigglob$ & 0.314 & 0.175 & 0.216 & 0.376 & 0.424 & 0.455 \\
 & $\viglay$ & 0.381 & 0.525 & 0.602 & 0.575 & 0.723 & 0.824 \\
\addlinespace[1pt]
General & $\Lreal$ & 1.614 & 1.407 & 1.240 & 1.191 & 1.095 & 1.012 \\
 & $\Lblank$ & 1.854 & 1.696 & 1.622 & 1.650 & 1.661 & 1.626 \\
 & $\vigglob$ & 0.192 & 0.235 & 0.296 & 0.354 & 0.436 & 0.448 \\
 & $\viglay$ & 0.048 & 0.054 & 0.087 & 0.104 & 0.129 & 0.166 \\
\addlinespace[1pt]
OCR & $\Lreal$ & 2.296 & 1.867 & 1.422 & 1.245 & 1.090 & 0.919 \\
 & $\Lblank$ & 2.457 & 2.187 & 1.972 & 1.878 & 1.804 & 1.741 \\
 & $\vigglob$ & 0.065 & 0.088 & 0.115 & 0.114 & 0.105 & 0.120 \\
 & $\viglay$ & 0.096 & 0.231 & 0.435 & 0.519 & 0.609 & 0.702 \\
\addlinespace[1pt]
STEM & $\Lreal$ & 1.713 & 1.377 & 1.110 & 0.988 & 0.881 & 0.779 \\
 & $\Lblank$ & 1.917 & 1.618 & 1.360 & 1.262 & 1.167 & 1.093 \\
 & $\vigglob$ & 0.136 & 0.151 & 0.130 & 0.141 & 0.124 & 0.137 \\
 & $\viglay$ & 0.068 & 0.090 & 0.120 & 0.133 & 0.162 & 0.177 \\
\addlinespace[1pt]
GUI & $\Lreal$ & 1.685 & 1.454 & 1.089 & 0.952 & 0.840 & 0.702 \\
 & $\Lblank$ & 1.820 & 1.677 & 1.547 & 1.515 & 1.495 & 1.459 \\
 & $\vigglob$ & 0.087 & 0.048 & 0.039 & 0.037 & 0.055 & 0.050 \\
 & $\viglay$ & 0.048 & 0.175 & 0.419 & 0.526 & 0.601 & 0.707 \\
\addlinespace[1pt]
\bottomrule
\end{tabular}
\end{table*}

\begin{table*}[p]\centering\small
\caption*{\textbf{158M, frozen SigLIP~2 encoder, 20B schedule.} Nats per text token, single checkpoints.}
\begin{tabular}{@{}llcccccc@{}}
\toprule
Image type & quantity & 1B & 2B & 5B & 10B & 20B & 40B \\
\midrule
Caption & $\Lreal$ & 3.246 & 2.872 & 2.650 & 2.514 & 2.324 & -- \\
 & $\Lblank$ & 4.387 & 4.019 & 3.902 & 3.803 & 3.620 & -- \\
 & $\vigglob$ & 0.645 & 0.609 & 0.627 & 0.620 & 0.620 & -- \\
 & $\viglay$ & 0.496 & 0.538 & 0.625 & 0.669 & 0.676 & -- \\
\addlinespace[1pt]
Chart & $\Lreal$ & 1.794 & 1.565 & 1.538 & 1.325 & 1.041 & -- \\
 & $\Lblank$ & 2.475 & 2.343 & 2.478 & 2.244 & 2.270 & -- \\
 & $\vigglob$ & 0.347 & 0.383 & 0.414 & 0.360 & 0.482 & -- \\
 & $\viglay$ & 0.335 & 0.394 & 0.525 & 0.559 & 0.748 & -- \\
\addlinespace[1pt]
General & $\Lreal$ & 1.557 & 1.381 & 1.279 & 1.242 & 1.103 & -- \\
 & $\Lblank$ & 1.835 & 1.696 & 1.717 & 1.702 & 1.673 & -- \\
 & $\vigglob$ & 0.231 & 0.266 & 0.365 & 0.354 & 0.424 & -- \\
 & $\viglay$ & 0.047 & 0.048 & 0.073 & 0.106 & 0.145 & -- \\
\addlinespace[1pt]
OCR & $\Lreal$ & 2.121 & 1.760 & 1.486 & 1.316 & 1.127 & -- \\
 & $\Lblank$ & 2.386 & 2.157 & 2.012 & 1.944 & 1.841 & -- \\
 & $\vigglob$ & 0.097 & 0.106 & 0.102 & 0.122 & 0.109 & -- \\
 & $\viglay$ & 0.168 & 0.291 & 0.424 & 0.506 & 0.604 & -- \\
\addlinespace[1pt]
STEM & $\Lreal$ & 1.612 & 1.329 & 1.155 & 1.048 & 0.909 & -- \\
 & $\Lblank$ & 1.820 & 1.547 & 1.403 & 1.322 & 1.210 & -- \\
 & $\vigglob$ & 0.129 & 0.130 & 0.134 & 0.141 & 0.147 & -- \\
 & $\viglay$ & 0.079 & 0.088 & 0.114 & 0.132 & 0.155 & -- \\
\addlinespace[1pt]
GUI & $\Lreal$ & 1.580 & 1.366 & 1.161 & 1.024 & 0.870 & -- \\
 & $\Lblank$ & 1.738 & 1.646 & 1.587 & 1.583 & 1.512 & -- \\
 & $\vigglob$ & 0.053 & 0.042 & 0.057 & 0.077 & 0.037 & -- \\
 & $\viglay$ & 0.105 & 0.238 & 0.369 & 0.483 & 0.605 & -- \\
\addlinespace[1pt]
\bottomrule
\end{tabular}
\end{table*}

\begin{table*}[p]\centering\small
\caption*{\textbf{158M, baseline seed 0, 20B schedule.} Nats per text token, single checkpoints.}
\begin{tabular}{@{}llcccccc@{}}
\toprule
Image type & quantity & 1B & 2B & 5B & 10B & 20B & 40B \\
\midrule
Caption & $\Lreal$ & 3.776 & 3.461 & 3.222 & 3.077 & 2.835 & -- \\
 & $\Lblank$ & 4.216 & 3.829 & 3.643 & 3.479 & 3.407 & -- \\
 & $\vigglob$ & 0.407 & 0.306 & 0.324 & 0.265 & 0.374 & -- \\
 & $\viglay$ & 0.034 & 0.062 & 0.097 & 0.137 & 0.198 & -- \\
\addlinespace[1pt]
Chart & $\Lreal$ & 2.204 & 1.999 & 1.867 & 1.652 & 1.342 & -- \\
 & $\Lblank$ & 2.541 & 2.304 & 2.278 & 2.094 & 1.940 & -- \\
 & $\vigglob$ & 0.271 & 0.147 & 0.162 & 0.270 & 0.225 & -- \\
 & $\viglay$ & 0.065 & 0.158 & 0.249 & 0.172 & 0.373 & -- \\
\addlinespace[1pt]
General & $\Lreal$ & 1.785 & 1.554 & 1.381 & 1.326 & 1.179 & -- \\
 & $\Lblank$ & 1.905 & 1.677 & 1.513 & 1.490 & 1.385 & -- \\
 & $\vigglob$ & 0.112 & 0.102 & 0.124 & 0.152 & 0.173 & -- \\
 & $\viglay$ & 0.008 & 0.021 & 0.008 & 0.013 & 0.033 & -- \\
\addlinespace[1pt]
OCR & $\Lreal$ & 2.270 & 2.032 & 1.875 & 1.770 & 1.642 & -- \\
 & $\Lblank$ & 2.305 & 2.057 & 1.906 & 1.829 & 1.731 & -- \\
 & $\vigglob$ & 0.033 & 0.020 & 0.016 & 0.033 & 0.048 & -- \\
 & $\viglay$ & 0.002 & 0.005 & 0.015 & 0.026 & 0.041 & -- \\
\addlinespace[1pt]
STEM & $\Lreal$ & 1.708 & 1.434 & 1.250 & 1.163 & 1.026 & -- \\
 & $\Lblank$ & 1.761 & 1.478 & 1.310 & 1.232 & 1.106 & -- \\
 & $\vigglob$ & 0.052 & 0.043 & 0.057 & 0.063 & 0.067 & -- \\
 & $\viglay$ & 0.000 & 0.001 & 0.003 & 0.005 & 0.013 & -- \\
\addlinespace[1pt]
GUI & $\Lreal$ & 1.707 & 1.609 & 1.539 & 1.505 & 1.451 & -- \\
 & $\Lblank$ & 1.734 & 1.623 & 1.566 & 1.534 & 1.484 & -- \\
 & $\vigglob$ & 0.027 & 0.014 & 0.022 & 0.019 & 0.007 & -- \\
 & $\viglay$ & 0.000 & 0.000 & 0.005 & 0.010 & 0.026 & -- \\
\addlinespace[1pt]
\bottomrule
\end{tabular}
\end{table*}

\begin{table*}[p]\centering\small
\caption*{\textbf{158M, baseline seed 1, 20B schedule.} Nats per text token, single checkpoints.}
\begin{tabular}{@{}llcccccc@{}}
\toprule
Image type & quantity & 1B & 2B & 5B & 10B & 20B & 40B \\
\midrule
Caption & $\Lreal$ & 3.803 & 3.459 & 3.201 & 3.074 & 2.850 & -- \\
 & $\Lblank$ & 4.212 & 3.818 & 3.586 & 3.559 & 3.469 & -- \\
 & $\vigglob$ & 0.381 & 0.302 & 0.299 & 0.373 & 0.454 & -- \\
 & $\viglay$ & 0.029 & 0.057 & 0.086 & 0.111 & 0.165 & -- \\
\addlinespace[1pt]
Chart & $\Lreal$ & 2.491 & 1.953 & 1.696 & 1.930 & 1.405 & -- \\
 & $\Lblank$ & 2.772 & 2.249 & 2.065 & 2.464 & 1.942 & -- \\
 & $\vigglob$ & 0.249 & 0.171 & 0.180 & 0.340 & 0.258 & -- \\
 & $\viglay$ & 0.032 & 0.125 & 0.190 & 0.194 & 0.279 & -- \\
\addlinespace[1pt]
General & $\Lreal$ & 2.032 & 1.597 & 1.424 & 1.314 & 1.202 & -- \\
 & $\Lblank$ & 2.130 & 1.749 & 1.528 & 1.581 & 1.534 & -- \\
 & $\vigglob$ & 0.097 & 0.146 & 0.102 & 0.257 & 0.294 & -- \\
 & $\viglay$ & 0.001 & 0.006 & 0.002 & 0.010 & 0.038 & -- \\
\addlinespace[1pt]
OCR & $\Lreal$ & 2.294 & 2.035 & 1.871 & 1.780 & 1.651 & -- \\
 & $\Lblank$ & 2.351 & 2.063 & 1.910 & 1.844 & 1.749 & -- \\
 & $\vigglob$ & 0.054 & 0.023 & 0.023 & 0.039 & 0.061 & -- \\
 & $\viglay$ & 0.003 & 0.005 & 0.015 & 0.026 & 0.037 & -- \\
\addlinespace[1pt]
STEM & $\Lreal$ & 1.746 & 1.441 & 1.268 & 1.145 & 1.021 & -- \\
 & $\Lblank$ & 1.830 & 1.485 & 1.316 & 1.231 & 1.112 & -- \\
 & $\vigglob$ & 0.085 & 0.043 & 0.043 & 0.079 & 0.078 & -- \\
 & $\viglay$ & $-$0.001 & 0.001 & 0.004 & 0.006 & 0.012 & -- \\
\addlinespace[1pt]
GUI & $\Lreal$ & 1.746 & 1.743 & 1.531 & 1.487 & 1.436 & -- \\
 & $\Lblank$ & 1.798 & 1.775 & 1.544 & 1.519 & 1.473 & -- \\
 & $\vigglob$ & 0.052 & 0.033 & 0.009 & 0.026 & 0.020 & -- \\
 & $\viglay$ & 0.000 & $-$0.001 & 0.004 & 0.006 & 0.016 & -- \\
\addlinespace[1pt]
\bottomrule
\end{tabular}
\end{table*}

\begin{table*}[p]\centering\small
\caption*{\textbf{158M, caption only, 20B schedule.} Nats per text token, single checkpoints.}
\begin{tabular}{@{}llcccccc@{}}
\toprule
Image type & quantity & 1B & 2B & 5B & 10B & 20B & 40B \\
\midrule
Caption & $\Lreal$ & 3.549 & 3.242 & 2.991 & 2.820 & 2.577 & -- \\
 & $\Lblank$ & 4.031 & 3.750 & 3.435 & 3.446 & 3.321 & -- \\
 & $\vigglob$ & 0.434 & 0.431 & 0.294 & 0.396 & 0.410 & -- \\
 & $\viglay$ & 0.048 & 0.077 & 0.150 & 0.230 & 0.334 & -- \\
\addlinespace[1pt]
Chart & $\Lreal$ & 3.902 & 3.670 & 3.565 & 3.711 & 3.757 & -- \\
 & $\Lblank$ & 4.096 & 3.645 & 3.491 & 3.654 & 3.723 & -- \\
 & $\vigglob$ & 0.199 & $-$0.022 & $-$0.093 & $-$0.079 & $-$0.047 & -- \\
 & $\viglay$ & $-$0.004 & $-$0.003 & 0.019 & 0.021 & 0.013 & -- \\
\addlinespace[1pt]
General & $\Lreal$ & 3.266 & 3.010 & 2.802 & 2.752 & 2.699 & -- \\
 & $\Lblank$ & 3.248 & 2.960 & 2.744 & 2.747 & 2.697 & -- \\
 & $\vigglob$ & $-$0.019 & $-$0.052 & $-$0.065 & $-$0.014 & $-$0.014 & -- \\
 & $\viglay$ & 0.001 & 0.002 & 0.007 & 0.010 & 0.012 & -- \\
\addlinespace[1pt]
OCR & $\Lreal$ & 3.051 & 2.708 & 2.525 & 2.447 & 2.348 & -- \\
 & $\Lblank$ & 3.234 & 2.738 & 2.525 & 2.442 & 2.362 & -- \\
 & $\vigglob$ & 0.175 & 0.023 & $-$0.007 & $-$0.010 & 0.016 & -- \\
 & $\viglay$ & 0.007 & 0.007 & 0.008 & 0.005 & $-$0.003 & -- \\
\addlinespace[1pt]
STEM & $\Lreal$ & 2.628 & 2.316 & 2.142 & 2.063 & 2.024 & -- \\
 & $\Lblank$ & 2.721 & 2.290 & 2.114 & 2.043 & 2.024 & -- \\
 & $\vigglob$ & 0.092 & $-$0.026 & $-$0.029 & $-$0.020 & $-$0.002 & -- \\
 & $\viglay$ & 0.000 & 0.000 & 0.001 & 0.001 & 0.002 & -- \\
\addlinespace[1pt]
GUI & $\Lreal$ & 3.021 & 2.754 & 2.606 & 2.508 & 2.444 & -- \\
 & $\Lblank$ & 3.224 & 2.788 & 2.614 & 2.539 & 2.471 & -- \\
 & $\vigglob$ & 0.203 & 0.033 & 0.008 & 0.029 & 0.027 & -- \\
 & $\viglay$ & 0.000 & 0.001 & 0.000 & 0.001 & 0.000 & -- \\
\addlinespace[1pt]
\bottomrule
\end{tabular}
\end{table*}

\begin{table*}[p]\centering\small
\caption*{\textbf{158M, document-heavy, 20B schedule.} Nats per text token, single checkpoints.}
\begin{tabular}{@{}llcccccc@{}}
\toprule
Image type & quantity & 1B & 2B & 5B & 10B & 20B & 40B \\
\midrule
Caption & $\Lreal$ & 4.109 & 3.744 & 3.498 & 3.364 & 3.158 & -- \\
 & $\Lblank$ & 4.581 & 4.118 & 3.866 & 3.726 & 3.639 & -- \\
 & $\vigglob$ & 0.459 & 0.348 & 0.306 & 0.299 & 0.391 & -- \\
 & $\viglay$ & 0.012 & 0.026 & 0.063 & 0.062 & 0.090 & -- \\
\addlinespace[1pt]
Chart & $\Lreal$ & 1.947 & 1.823 & 1.621 & 1.542 & 1.298 & -- \\
 & $\Lblank$ & 2.431 & 2.124 & 2.053 & 2.060 & 1.906 & -- \\
 & $\vigglob$ & 0.433 & 0.158 & 0.222 & 0.257 & 0.237 & -- \\
 & $\viglay$ & 0.051 & 0.143 & 0.209 & 0.262 & 0.372 & -- \\
\addlinespace[1pt]
General & $\Lreal$ & 1.618 & 1.456 & 1.356 & 1.307 & 1.183 & -- \\
 & $\Lblank$ & 1.688 & 1.537 & 1.456 & 1.440 & 1.365 & -- \\
 & $\vigglob$ & 0.067 & 0.076 & 0.089 & 0.130 & 0.168 & -- \\
 & $\viglay$ & 0.003 & 0.004 & 0.011 & 0.003 & 0.014 & -- \\
\addlinespace[1pt]
OCR & $\Lreal$ & 2.193 & 1.952 & 1.797 & 1.704 & 1.586 & -- \\
 & $\Lblank$ & 2.233 & 1.982 & 1.847 & 1.775 & 1.698 & -- \\
 & $\vigglob$ & 0.037 & 0.023 & 0.031 & 0.039 & 0.069 & -- \\
 & $\viglay$ & 0.003 & 0.007 & 0.019 & 0.032 & 0.043 & -- \\
\addlinespace[1pt]
STEM & $\Lreal$ & 1.616 & 1.356 & 1.193 & 1.102 & 0.984 & -- \\
 & $\Lblank$ & 1.683 & 1.406 & 1.249 & 1.168 & 1.064 & -- \\
 & $\vigglob$ & 0.067 & 0.049 & 0.053 & 0.057 & 0.069 & -- \\
 & $\viglay$ & $-$0.001 & 0.002 & 0.003 & 0.008 & 0.012 & -- \\
\addlinespace[1pt]
GUI & $\Lreal$ & 1.712 & 1.614 & 1.539 & 1.495 & 1.438 & -- \\
 & $\Lblank$ & 1.736 & 1.636 & 1.572 & 1.521 & 1.467 & -- \\
 & $\vigglob$ & 0.022 & 0.022 & 0.028 & 0.016 & 0.011 & -- \\
 & $\viglay$ & 0.002 & 0.000 & 0.005 & 0.010 & 0.018 & -- \\
\addlinespace[1pt]
\bottomrule
\end{tabular}
\end{table*}

\begin{table*}[p]\centering\small
\caption*{\textbf{158M, 16-pixel patches, 20B schedule.} Nats per text token, single checkpoints.}
\begin{tabular}{@{}llcccccc@{}}
\toprule
Image type & quantity & 1B & 2B & 5B & 10B & 20B & 40B \\
\midrule
Caption & $\Lreal$ & 4.115 & 3.759 & 3.508 & 3.336 & 3.132 & -- \\
 & $\Lblank$ & 4.826 & 4.126 & 3.884 & 3.756 & 3.628 & -- \\
 & $\vigglob$ & 0.700 & 0.344 & 0.318 & 0.335 & 0.391 & -- \\
 & $\viglay$ & 0.011 & 0.023 & 0.058 & 0.085 & 0.105 & -- \\
\addlinespace[1pt]
Chart & $\Lreal$ & 2.484 & 2.045 & 1.796 & 1.734 & 1.614 & -- \\
 & $\Lblank$ & 3.003 & 2.336 & 2.245 & 2.159 & 2.092 & -- \\
 & $\vigglob$ & 0.463 & 0.240 & 0.274 & 0.245 & 0.213 & -- \\
 & $\viglay$ & 0.056 & 0.051 & 0.175 & 0.180 & 0.265 & -- \\
\addlinespace[1pt]
General & $\Lreal$ & 1.837 & 1.591 & 1.456 & 1.455 & 1.296 & -- \\
 & $\Lblank$ & 2.312 & 1.846 & 1.677 & 1.849 & 1.725 & -- \\
 & $\vigglob$ & 0.457 & 0.252 & 0.220 & 0.285 & 0.377 & -- \\
 & $\viglay$ & 0.017 & 0.003 & 0.001 & 0.109 & 0.052 & -- \\
\addlinespace[1pt]
OCR & $\Lreal$ & 2.474 & 2.195 & 2.010 & 1.900 & 1.765 & -- \\
 & $\Lblank$ & 2.732 & 2.236 & 2.048 & 1.942 & 1.828 & -- \\
 & $\vigglob$ & 0.257 & 0.038 & 0.035 & 0.036 & 0.050 & -- \\
 & $\viglay$ & 0.001 & 0.003 & 0.003 & 0.006 & 0.013 & -- \\
\addlinespace[1pt]
STEM & $\Lreal$ & 1.954 & 1.605 & 1.406 & 1.286 & 1.149 & -- \\
 & $\Lblank$ & 2.196 & 1.659 & 1.467 & 1.354 & 1.229 & -- \\
 & $\vigglob$ & 0.244 & 0.054 & 0.060 & 0.060 & 0.072 & -- \\
 & $\viglay$ & $-$0.002 & 0.000 & 0.001 & 0.008 & 0.009 & -- \\
\addlinespace[1pt]
GUI & $\Lreal$ & 1.777 & 1.653 & 1.587 & 1.550 & 1.485 & -- \\
 & $\Lblank$ & 1.887 & 1.662 & 1.608 & 1.619 & 1.519 & -- \\
 & $\vigglob$ & 0.110 & 0.011 & 0.021 & 0.069 & 0.030 & -- \\
 & $\viglay$ & 0.000 & $-$0.001 & 0.000 & 0.001 & 0.004 & -- \\
\addlinespace[1pt]
\bottomrule
\end{tabular}
\end{table*}

\begin{table*}[p]\centering\small
\caption*{\textbf{158M, random-init ViT, stopped at 8B tokens.} Nats per text token, single checkpoints.}
\begin{tabular}{@{}llcccccc@{}}
\toprule
Image type & quantity & 1B & 2B & 5B & 10B & 20B & 40B \\
\midrule
Caption & $\Lreal$ & 3.819 & 3.490 & 3.233 & -- & -- & -- \\
 & $\Lblank$ & 3.929 & 3.652 & 3.481 & -- & -- & -- \\
 & $\vigglob$ & 0.110 & 0.146 & 0.174 & -- & -- & -- \\
 & $\viglay$ & 0.000 & 0.016 & 0.074 & -- & -- & -- \\
\addlinespace[1pt]
Chart & $\Lreal$ & 2.002 & 1.809 & 1.735 & -- & -- & -- \\
 & $\Lblank$ & 2.027 & 1.847 & 1.913 & -- & -- & -- \\
 & $\vigglob$ & 0.025 & 0.032 & 0.125 & -- & -- & -- \\
 & $\viglay$ & 0.000 & 0.006 & 0.053 & -- & -- & -- \\
\addlinespace[1pt]
General & $\Lreal$ & 1.587 & 1.436 & 1.328 & -- & -- & -- \\
 & $\Lblank$ & 1.603 & 1.470 & 1.395 & -- & -- & -- \\
 & $\vigglob$ & 0.016 & 0.034 & 0.061 & -- & -- & -- \\
 & $\viglay$ & 0.000 & 0.000 & 0.007 & -- & -- & -- \\
\addlinespace[1pt]
OCR & $\Lreal$ & 2.274 & 2.037 & 1.845 & -- & -- & -- \\
 & $\Lblank$ & 2.282 & 2.044 & 1.866 & -- & -- & -- \\
 & $\vigglob$ & 0.007 & 0.007 & 0.017 & -- & -- & -- \\
 & $\viglay$ & 0.000 & 0.000 & 0.005 & -- & -- & -- \\
\addlinespace[1pt]
STEM & $\Lreal$ & 1.669 & 1.415 & 1.229 & -- & -- & -- \\
 & $\Lblank$ & 1.679 & 1.436 & 1.279 & -- & -- & -- \\
 & $\vigglob$ & 0.011 & 0.021 & 0.047 & -- & -- & -- \\
 & $\viglay$ & 0.000 & 0.000 & 0.003 & -- & -- & -- \\
\addlinespace[1pt]
GUI & $\Lreal$ & 1.722 & 1.613 & 1.529 & -- & -- & -- \\
 & $\Lblank$ & 1.723 & 1.623 & 1.541 & -- & -- & -- \\
 & $\vigglob$ & 0.001 & 0.009 & 0.012 & -- & -- & -- \\
 & $\viglay$ & 0.000 & 0.001 & 0.000 & -- & -- & -- \\
\addlinespace[1pt]
\bottomrule
\end{tabular}
\end{table*}

\begin{table*}[p]\centering\small
\caption*{\textbf{158M, batch 512, 40B schedule.} Nats per text token, single checkpoints.}
\begin{tabular}{@{}llcccccc@{}}
\toprule
Image type & quantity & 1B & 2B & 5B & 10B & 20B & 40B \\
\midrule
Caption & $\Lreal$ & 3.998 & 3.578 & 3.260 & 3.091 & 2.923 & 2.763 \\
 & $\Lblank$ & 4.523 & 4.061 & 3.958 & 3.739 & 3.542 & 3.497 \\
 & $\vigglob$ & 0.516 & 0.445 & 0.625 & 0.550 & 0.483 & 0.540 \\
 & $\viglay$ & 0.008 & 0.038 & 0.073 & 0.098 & 0.137 & 0.195 \\
\addlinespace[1pt]
Chart & $\Lreal$ & 2.207 & 1.891 & 1.759 & 1.580 & 1.373 & 1.274 \\
 & $\Lblank$ & 2.660 & 2.235 & 2.217 & 2.064 & 1.924 & 1.888 \\
 & $\vigglob$ & 0.441 & 0.247 & 0.278 & 0.310 & 0.301 & 0.255 \\
 & $\viglay$ & 0.013 & 0.098 & 0.180 & 0.174 & 0.250 & 0.359 \\
\addlinespace[1pt]
General & $\Lreal$ & 1.705 & 1.486 & 1.346 & 1.306 & 1.211 & 1.138 \\
 & $\Lblank$ & 1.825 & 1.575 & 1.497 & 1.454 & 1.423 & 1.374 \\
 & $\vigglob$ & 0.120 & 0.089 & 0.147 & 0.141 & 0.190 & 0.187 \\
 & $\viglay$ & 0.001 & 0.001 & 0.004 & 0.008 & 0.022 & 0.050 \\
\addlinespace[1pt]
OCR & $\Lreal$ & 2.492 & 2.124 & 1.889 & 1.781 & 1.689 & 1.600 \\
 & $\Lblank$ & 2.543 & 2.154 & 1.925 & 1.832 & 1.768 & 1.694 \\
 & $\vigglob$ & 0.049 & 0.028 & 0.027 & 0.034 & 0.054 & 0.056 \\
 & $\viglay$ & 0.001 & 0.001 & 0.008 & 0.017 & 0.026 & 0.038 \\
\addlinespace[1pt]
STEM & $\Lreal$ & 1.946 & 1.502 & 1.267 & 1.162 & 1.056 & 0.976 \\
 & $\Lblank$ & 2.027 & 1.565 & 1.329 & 1.232 & 1.138 & 1.062 \\
 & $\vigglob$ & 0.081 & 0.064 & 0.059 & 0.066 & 0.075 & 0.075 \\
 & $\viglay$ & 0.000 & 0.000 & 0.003 & 0.003 & 0.007 & 0.011 \\
\addlinespace[1pt]
GUI & $\Lreal$ & 1.795 & 1.655 & 1.534 & 1.497 & 1.457 & 1.412 \\
 & $\Lblank$ & 1.823 & 1.690 & 1.550 & 1.529 & 1.488 & 1.450 \\
 & $\vigglob$ & 0.028 & 0.034 & 0.016 & 0.026 & 0.017 & 0.011 \\
 & $\viglay$ & 0.000 & 0.000 & 0.000 & 0.006 & 0.014 & 0.026 \\
\addlinespace[1pt]
\bottomrule
\end{tabular}
\end{table*}

\begin{table*}[p]\centering\small
\caption*{\textbf{158M, batch 32, 20B schedule.} Nats per text token, single checkpoints.}
\begin{tabular}{@{}llcccccc@{}}
\toprule
Image type & quantity & 1B & 2B & 5B & 10B & 20B & 40B \\
\midrule
Caption & $\Lreal$ & 3.805 & 3.569 & 3.404 & 3.342 & 3.011 & -- \\
 & $\Lblank$ & 4.092 & 3.926 & 3.783 & 3.681 & 3.494 & -- \\
 & $\vigglob$ & 0.244 & 0.297 & 0.291 & 0.226 & 0.291 & -- \\
 & $\viglay$ & 0.042 & 0.060 & 0.088 & 0.113 & 0.193 & -- \\
\addlinespace[1pt]
Chart & $\Lreal$ & 2.205 & 2.075 & 2.070 & 1.890 & 1.516 & -- \\
 & $\Lblank$ & 2.456 & 2.524 & 2.386 & 2.266 & 1.989 & -- \\
 & $\vigglob$ & 0.183 & 0.272 & 0.112 & 0.211 & 0.109 & -- \\
 & $\viglay$ & 0.068 & 0.177 & 0.205 & 0.165 & 0.365 & -- \\
\addlinespace[1pt]
General & $\Lreal$ & 1.779 & 1.630 & 1.473 & 1.443 & 1.257 & -- \\
 & $\Lblank$ & 1.917 & 1.828 & 1.656 & 1.727 & 1.561 & -- \\
 & $\vigglob$ & 0.108 & 0.094 & 0.160 & 0.272 & 0.269 & -- \\
 & $\viglay$ & 0.030 & 0.104 & 0.023 & 0.012 & 0.035 & -- \\
\addlinespace[1pt]
OCR & $\Lreal$ & 2.287 & 2.104 & 2.012 & 1.970 & 1.756 & -- \\
 & $\Lblank$ & 2.307 & 2.135 & 2.048 & 2.022 & 1.823 & -- \\
 & $\vigglob$ & 0.017 & 0.024 & 0.026 & 0.035 & 0.029 & -- \\
 & $\viglay$ & 0.003 & 0.008 & 0.011 & 0.017 & 0.038 & -- \\
\addlinespace[1pt]
STEM & $\Lreal$ & 1.723 & 1.514 & 1.391 & 1.372 & 1.126 & -- \\
 & $\Lblank$ & 1.752 & 1.564 & 1.438 & 1.435 & 1.197 & -- \\
 & $\vigglob$ & 0.029 & 0.047 & 0.042 & 0.060 & 0.059 & -- \\
 & $\viglay$ & 0.000 & 0.003 & 0.005 & 0.003 & 0.012 & -- \\
\addlinespace[1pt]
GUI & $\Lreal$ & 1.722 & 1.646 & 1.604 & 1.612 & 1.529 & -- \\
 & $\Lblank$ & 1.752 & 1.673 & 1.622 & 1.635 & 1.565 & -- \\
 & $\vigglob$ & 0.030 & 0.027 & 0.009 & 0.013 & 0.001 & -- \\
 & $\viglay$ & 0.000 & 0.001 & 0.009 & 0.010 & 0.036 & -- \\
\addlinespace[1pt]
\bottomrule
\end{tabular}
\end{table*}

\clearpage

\end{document}